\documentclass[10pt,twocolumn,letterpaper]{article}

\usepackage{cvpr}              % To produce the CAMERA-READY version
\definecolor{cvprblue}{rgb}{0.21,0.49,0.74}
\usepackage[pagebackref,breaklinks,colorlinks,allcolors=cvprblue]{hyperref}
\usepackage{multirow}
\usepackage{dblfloatfix}
\usepackage{color}

\def\paperID{3041} % *** Enter the Paper ID here
\def\confName{CVPR}
\def\confYear{2025}

\title{Synthetic-to-Real Self-supervised Robust Depth Estimation\\ via Learning with Motion and Structure Priors}

\author{
\vspace{-8mm}
Weilong Yan$^{1}$ \vspace{-4.3mm}\\
\and
Ming Li$^{3}$ \vspace{-4.3mm}\\
\and
Haipeng Li$^{4}$ \vspace{-4.3mm}\\
\and
Shuwei Shao$^{5}$ \vspace{-4.3mm}\\
\and
Robby T. Tan$^{1,2}$ \vspace{-4.3mm}\\
\and
\textsuperscript{1}National University of Singapore \hspace{10mm} 
\textsuperscript{2}ASUS Intelligent Cloud Services \\
\textsuperscript{3}Guangdong Laboratory of Artificial Intelligence and Digital Economy (SZ) \\
\textsuperscript{4}University of Electronic Science and Technology of China \\
\textsuperscript{5}School of Control Science and Engineering, Shandong University \\
\vspace{-4mm}
\centering
{\tt\small yanweilong@u.nus.edu}
}

\begin{document}
\maketitle

\begin{abstract}
Self-supervised depth estimation from monocular cameras in diverse outdoor conditions, such as daytime, rain, and nighttime, is challenging due to the difficulty of learning universal representations and the severe lack of labeled real-world adverse data.
Previous methods either rely on synthetic inputs and pseudo-depth labels or directly apply daytime strategies to adverse conditions, resulting in suboptimal results.
In this paper, we present the first synthetic-to-real robust depth estimation framework, incorporating motion and structure priors to capture real-world knowledge effectively. 
In the synthetic adaptation, we transfer motion-structure knowledge inside cost volumes for better robust representation, using a frozen daytime model to train a depth estimator in synthetic adverse conditions.
In the innovative real adaptation, which targets to fix synthetic-real gaps, models trained earlier identify the weather-insensitive regions with a designed consistency-reweighting strategy to emphasize valid pseudo-labels.
We introduce a new regularization by gathering explicit depth distribution to constrain the model facing real-world data.
Experiments show that our method outperforms the state-of-the-art across diverse conditions in multi-frame and single-frame evaluations. We achieve improvements of 7.5\% and 4.3\% in AbsRel and RMSE on average for nuScenes and Robotcar datasets (daytime, nighttime, rain). In zero-shot evaluation of DrivingStereo (rain, fog), our method generalizes better than previous ones. The code is at 
\href{https://github.com/DavidYan2001/Synthetic2Real-Depth}{Syn2Real-Depth}.

\end{abstract}

\section{Introduction}
\label{sec:intro}

% \documentclass{article}
% \usepackage{graphicx}
% \usepackage{subcaption}
% \usepackage{array}
% \usepackage[table,xcdraw]{xcolor}
% \usepackage{amsmath}

% \begin{figure}[htb]
% \centering
% \includegraphics[width=1.\linewidth, height=.4\linewidth]{figures/figure_idea_v4.png}
% % \vspace{-0.1cm}
% \caption{Illustration of the main idea. }
% \label{fig:idea}
% \vspace{-0.4cm}
% \end{figure}

\begin{figure}[htb]
\centering
\includegraphics[width=1.\linewidth, height=.4\linewidth]
{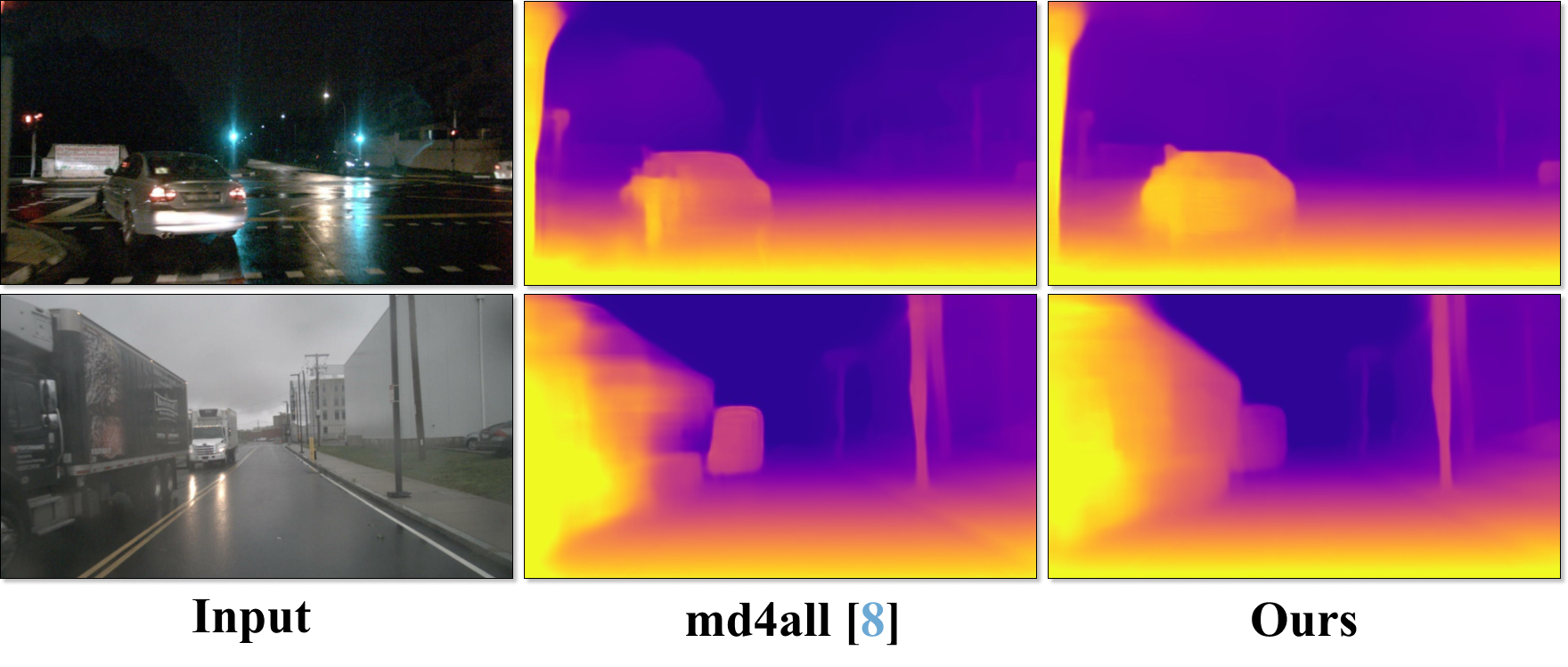}
% \vspace{-0.1cm}
\caption{Visual comparison between the previous \cite{gasperini_morbitzer2023md4all} and ours. The first and second rows correspond to nighttime and rain, where ours provides more robust estimations especially for the vehicles. }
\vspace{-0.4cm}
\label{fig:teaser}
\vspace{-0.1cm}
\end{figure}

% Depth estimation from monocular cameras is a fundamental task in computer vision and has advanced lots of downstream tasks, including autonomous driving, robotics and augmented reality. 
% %
% Due to the high cost and difficulty of obtaining reliable depth ground truth, self-supervised methods which focus on photometric and geometric consistency \cite{godard2019digging, zhao2022monovit, wang2023sqldepth, bian2021ijcv_scdepth, 2023planedepth, 2021epcdepth, litemono} have been explored to reduce the need for expensive hardware. 
% %
% Early self-supervised methods take a single image for depth inference, while recently several multi-frame based methods \cite{manydepth, disentangling, multi_frame_transformer, bangunharcana2023dualrefine, prodepth} for monocular videos have been proposed to achieve more accurate results. 
% %
% However, these approaches all make assumption of clear scenes such as daytime, which struggles to generalize to diverse adverse conditions.

Depth estimation from monocular cameras is a fundamental task in computer vision, essential for applications such as autonomous driving, robotics, and closely connected with segmentation \cite{xin2024heap}. Due to the high cost and difficulty of obtaining reliable ground truth, self-supervised methods focusing on photometric and geometric consistency \cite{2017cvpr_egomotion,godard2019digging, zhao2022monovit, wang2023sqldepth, bian2021ijcv_scdepth, 2023planedepth, 2021epcdepth, litemono} have been developed to reduce dependence on expensive hardware. Early self-supervised methods perform depth inference on single images \cite{2017cvpr_egomotion,godard2019digging}, while recent multi-frame methods \cite{manydepth, disentangling, multi_frame_transformer, bangunharcana2023dualrefine, prodepth} for monocular videos achieve improved accuracy. However, these methods generally assume daytime conditions, limiting their generalizability to various adverse scenarios.

Predicting depth in adverse conditions within a single model faces two major challenges: $(1)$ the diverse types of degradation make it difficult to learn a unified representation; $(2)$ the lack of labeled real data makes it difficult to learn in adverse conditions in the real world.
%
% insufficient exploitation of unlabeled real-world data limits the model's effectiveness in real adverse environments.
%
To overcome these challenges, many methods are designed for depth estimation in adverse conditions within an unsupervised domain adaptive (UDA) manner \cite{diffusionadverse, gasperini_morbitzer2023md4all, Saunders_2023_ICCV, wang2023weatherdepth, diffusion_contrast}. 
They aim to adapt knowledge from daytime to different adverse conditions and can be divided into single-condition and multi-condition categories.
Single-condition methods \cite{vankadari2020ADFA, wang2021RNW, spencer2020defeat, 23ICRA_Steps} focus on a particular case (e.g., nighttime) with specific designs, but are quite limited to work under multiple conditions in real-world applications.
%
% Some methods \cite{liu2021ADIDS, vankadari2023sun} target at daytime and nighttime but perform poor in both domains and still can not generalize to multiple conditions.
%
Some methods~\cite{liu2021ADIDS, vankadari2023sun} aim to address both daytime and nighttime conditions but exhibit limited performance across these domains, often failing to generalize to multiple conditions.

% Recently, several approaches \cite{gasperini_morbitzer2023md4all, Saunders_2023_ICCV, diffusionadverse, wang2023weatherdepth, diffusion_contrast} emerged to obtain a single model to work well in diverse conditions. 
% %
% They either explore data augmentation and generation techniques \cite{CycleGAN2017, zheng_2020_forkgan, LDM, controlnet} to create paired data and utilize a daytime model for teacher-student training, or directly apply the daytime-used photometric loss to adverse conditions and a generative adversarial network (GAN) to discriminate the predicted depth.
% %
% Although thsese methods advanced the field of robust depth estimation a lot, the challenges of degradation and real data utilization are not solved significantly.
% %
% On one hand, learning with only depth pseudo-labels is not enough to obtain a common representation due to the heterogeneous nature of different conditions \cite{diffusion_contrast, zhou2024exploring}, which inspires us to find an auxiliary space for better representation.
% %
% On the other hand, applying GAN or photometric loss for real adverse data either provides implicit constraint to the prediction with randomness, or violates the brightness consistency assumption \cite{gasperini_morbitzer2023md4all, Saunders_2023_ICCV};
% %
% and only utilizing augmented and synthetic data will drive the model to overfit to the synthetic domain, which still has large gaps compared with the real domain. 
% %
% This motivates us to seek for better strategy to learn from real data.
Recently, several approaches \cite{gasperini_morbitzer2023md4all, Saunders_2023_ICCV, diffusionadverse, wang2023weatherdepth, diffusion_contrast} have emerged to develop a single model that performs well across diverse conditions. These methods either explore data augmentation and generation techniques \cite{CycleGAN2017, zheng_2020_forkgan, LDM, controlnet, Wu2024TowardsBT} to create paired data and employ teacher-student training, or apply photometric loss directly to adverse conditions, alongside a generative adversarial network (GAN) to discriminate the style of predicted depth.
While these approaches have advanced the community, they still face challenges in learning robustness facing real-world degradation.
On one hand, solely learning from depth pseudo-labels is insufficient for robust common representation due to the heterogeneous nature of different conditions \cite{diffusion_contrast, zhou2024exploring}.
On the other hand, applying GANs or photometric loss to real adverse data either imposes implicit constraints with sampling randomness or violates the brightness consistency assumption \cite{gasperini_morbitzer2023md4all, Saunders_2023_ICCV}. 
Furthermore, relying only on synthetic data risks overfitting to the synthetic domain, which still faces the out-of-distribution issue.
These motivate us to explore an auxiliary space for improved representation learning and an effective strategy to learn in real-world conditions.

% In this work, we target to find answers to the mentioned challenges and the weaknesses of existing methods. 
% %
% The main idea of our proposed method is illustrated in Fig.\ref{fig:idea} (b).
% %
% Compared with previous methods in Fig.\ref{fig:idea} (a), we are the first to propose synthetic-to-real monocular robust depth estimation in diverse conditions with multi-frames or single-frame against degradation. 
% %
% After obtaining a strong self-supervised daytime model, we transfer the knowledge to diverse adverse conditions.
% %
% Firstly, inspired by \cite{manydepth, disentangling, prodepth}, we find that motion with the cost volume is a proper auxiliary space for robust representation learning compared with feature distillation, due to its intrinsic property of filtering out redundant information and retaining motion and structure characters.
% %
% Thus, in the synthetic adaptation ($\textbf{SA}$), apart from supervision with depth pseudo-labels, we build up cost volumes on the high-level features with the learned pose information, and transfer daytime motion-spatial knowledge via the cost volumes to adverse conditions.
%
In this work, we propose the first synthetic-to-real framework to achieve depth estimation under diverse conditions. Different from previous methods focused solely on synthetic data, ours learns from the synthetic to real-world data. 
Since only pseudo-depth label is insufficient for the model to learn due to the heterogeneous features \cite{diffusion_contrast, zhou2024exploring}, we dig into motion constraint and explicit structure prior for efficient learning of real-world knowledge. 
Examples of our method compared with previous ones are shown in Fig.\ref{fig:teaser}.
%
% Compared to previous methods in Fig.\ref{fig:idea} (a), we are the first to propose synthetic-to-real monocular robust depth estimation under diverse conditions, leveraging multi-frame and single-frame inputs to fight against degradation.
%
Starting from having a strong self-supervised daytime model, we identify the cost volume \cite{manydepth, disentangling, prodepth} as an effective auxiliary space for robust representation, as it naturally filters redundant information while preserving motion and structural features.
Thus, in the synthetic adaptation (\textbf{SA}) which learns on the synthetic data, we construct cost volumes from high-level features to transfer filtered daytime motion-spatial knowledge to adverse conditions, apart from learning only with pseudo-depth labels.

% Moreover, to fix the synthetic-real gaps, we elaborately design a real adaptation $(\textbf{RA})$ stage.
% %
% Specifically, the pretrained daytime model and the synthetic model in SA are leveraged together to give a consistency map for real data of multiple conditions, which will be utilized in a consistency-reweighting strategy to emphasize the valid parts of pseudo-labels. 
% %
% Since the pseudo-labels from the synthetic model still owns problems with the synthetic-real gap, we dig into the explicit priors of daytime and depth information, by calculating the predicted depth distribution from the daytime model. By analyzing the similarity and difference of predicted depth distributions between daytime and real adverse conditions, we set the daytime distribution as structure priors which will constrain predictions of the real adverse data.
To address the synthetic-real gaps, we propose a novel real adaptation $(\textbf{RA})$ stage. One challenge is to find valid supervision signals from earlier trained models, and we design a consistency-reweighting strategy by incorporating different models to find a consistency map that can emphasize the valid parts of pseudo-labels and provide effective signals for learning in real-world conditions.
Since pseudo-labels from the frozen model still suffer from the synthetic-real gap, we introduce an effective prior based on daytime depth predictions, by calculating explicit depth distribution from the daytime model and setting it to be a reference in diverse conditions. This structural prior provides explicit constraint when facing real adverse data, learning real-world  information further effectively.

With the proposed synthetic-to-real adaptation pipeline, our model can provide much more robust depth predictions in diverse conditions through different real-world datasets \cite{nuscenes2019, robotcar, drivingstereo}. Our contributions are listed as follows:
\begin{itemize}
  \item We propose a novel synthetic-to-real adaptation pipeline for robust depth estimation in diverse conditions, which is able to work in multiple real-world datasets. 
  \item We construct the motion and cost volume as auxiliary space to transfer filtered daytime knowledge for depth, which boosts the robust representation learning.
  \item We design a novel real adaptation with two strategies named consistency-reweighting and structure prior constraint, which effectively learns from real adverse data.
  \item Through the experiments, our method achieves state-of-the-art performance on multiple real challenging datasets (improving by $7.5\%$ in AbsRel and $4.3\%$ in RMSE), in multi-frame, single-frame, zero-shot settings.
\end{itemize}

%-------------------------------------------------------------------------

% \begin{figure*}
%   \centering
%   \begin{subfigure}{0.68\linewidth}
%     \fbox{\rule{0pt}{2in} \rule{.9\linewidth}{0pt}}
%     \caption{An example of a subfigure.}
%     \label{fig:short-a}
%   \end{subfigure}
%   \hfill
%   \begin{subfigure}{0.28\linewidth}
%     \fbox{\rule{0pt}{2in} \rule{.9\linewidth}{0pt}}
%     \caption{Another example of a subfigure.}
%     \label{fig:short-b}
%   \end{subfigure}
%   \caption{Example of a short caption, which should be centered.}
%   \label{fig:short}
% \end{figure*}

\section{Related Work}
\label{sec:related_works}

%-------------------------------------------------------------------------
\subsection{Self-supervised Depth Estimation}
% Self-supervised depth estimation has been explored constantly for its requirement of only RGB data. Previously, SfM-Learner \cite{2017cvpr_egomotion} is designed based on the photometric consistency to turn the depth estimation into a view reconstruction task. After that, Monodepth2 \cite{godard2019digging} introduces per-pixel photometric loss and auto-masking strategy to reduce the influence of moving objects. SC-DepthV1 \cite{bian2021ijcv_scdepth} and V3 \cite{sc_depthv3} target to solve depth scale ambiguity and high dynamic scene, respectively. MonoViT \cite{zhao2022monovit} combines the abilities of CNNs and Transformers, while Lite-mono \cite{litemono} proposes a lightweight pipeline. These single-frame methods inspire multi-frame design. Manydepth \cite{manydepth} was proposed to leverage cost volume built on consecutive frames to achieve better accuracy, DepthFormer \cite{multi_frame_transformer} designed an attention mechanism to refine the cost volume, DualRefine \cite{bangunharcana2023dualrefine} proposes to iteratively update the depth and pose, DynamicDepth \cite{disentangling} and Prodepth \cite{prodepth} design different strategies to solve dynamic problems. Although these methods work quite well in daytime, they usually suffer from adverse conditions.
In the deep learning era \cite{Liming_2021_ICCV, li2023instant3d, LodoNet, LiMingTMM2021, Li_2023_ICCV}, self-supervised depth has been widely studied for its reliance on only RGB data. SfM-Learner \cite{2017cvpr_egomotion} firstly leveraged photometric consistency to reformulate depth estimation as a view reconstruction task. Monodepth2 \cite{godard2019digging} introduced per-pixel photometric loss and auto-masking to reduce the impact of moving objects. SC-DepthV1 \cite{bian2021ijcv_scdepth} and V3 \cite{sc_depthv3} aimed to address depth scale ambiguity and high dynamic scenes, respectively. MonoViT \cite{zhao2022monovit} combines CNN and Transformer capabilities, while Lite-mono \cite{litemono} offers a lightweight approach. These single-frame methods inspired multi-frame designs, such as Manydepth \cite{manydepth}, which utilizes cost volumes across consecutive frames to improve accuracy. DepthFormer \cite{multi_frame_transformer} introduced an attention mechanism for refining cost volumes, DualRefine \cite{bangunharcana2023dualrefine} iteratively updates depth and pose, while DynamicDepth \cite{disentangling} and Prodepth \cite{prodepth} tackle dynamic object challenges. Though effective in daytime, these methods typically struggle in adverse conditions.

\begin{figure*}[!t]
\centering
\includegraphics[width=1.\linewidth, height=.57\linewidth]{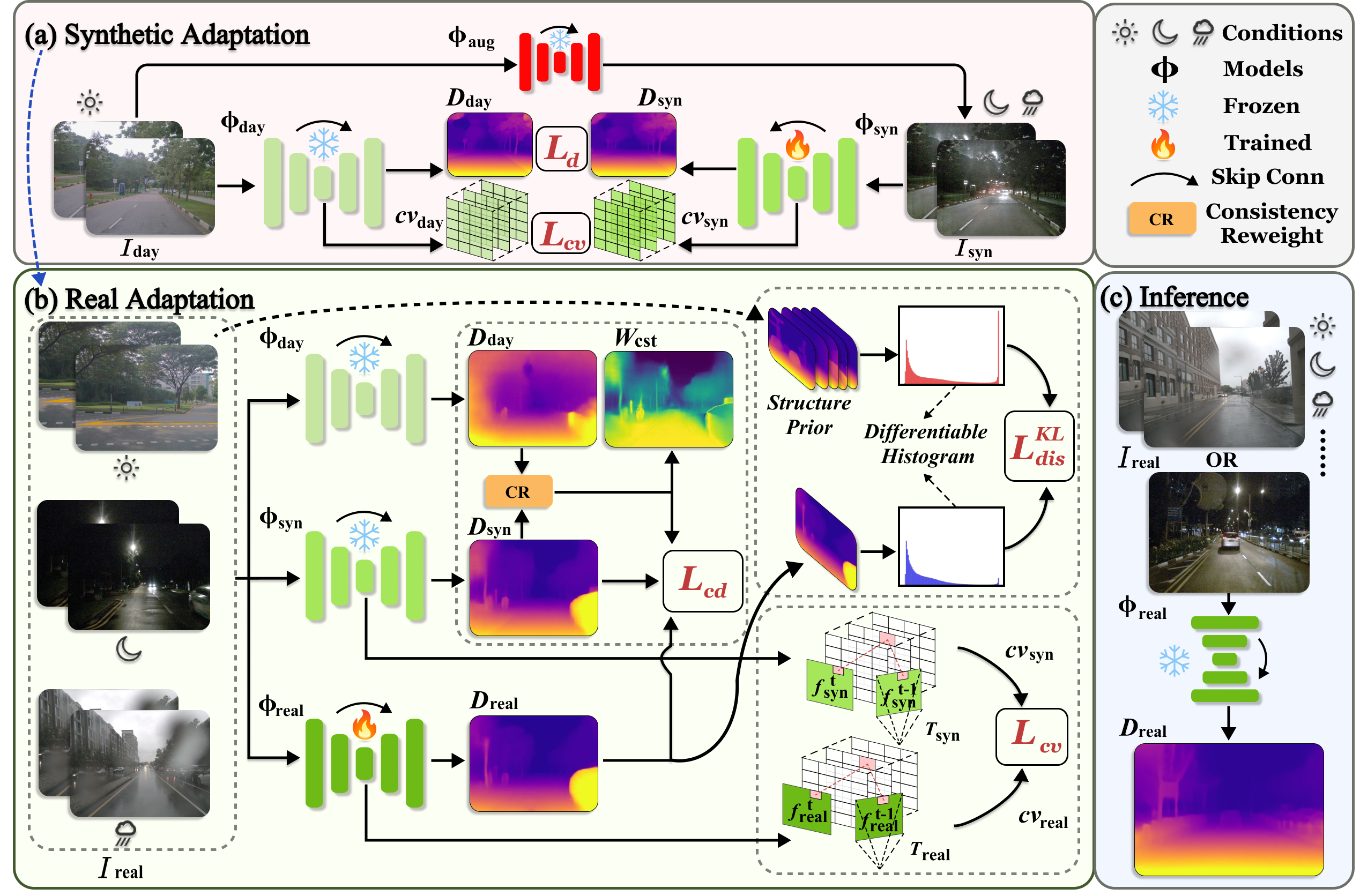}
% \vspace{-0.1cm}
\caption{Illustration of our proposed pipeline. (a) Synthetic adaptation utilizes an augmentation model to generate paired data for training, and we also conduct learning in an auxiliary motion space. (b) Real adaptation leverages the daytime model and synthetic model to provide valid pseudo-labels with consistency reweighting, and the daytime depth distribution from plenty of daytime predictions are seen as  structure prior to constrain the model facing real adverse data. (c) Inference stage can work in multi-frame and single-frame settings.}
\label{fig:pipeline}
\vspace{-0.3cm}
\end{figure*}
\subsection{Robust Depth Estimation}

Early approaches are tailored to specific conditions. ADFA \cite{vankadari2020ADFA} and RNW \cite{wang2021RNW} use GANs to distinguish features and depth maps in nighttime conditions, while ITDFA \cite{zhao2022ITDFA} employs GANs to translate daytime images to nighttime. STEPS \cite{23ICRA_Steps} and RNW \cite{wang2021RNW} focus on image enhancement to aid depth estimation. However, these methods, with some works \cite{spencer2020defeat, liu2021ADIDS, vankadari2023sun} addressing daytime and nighttime, are limited in real-world applications of multiple conditions.

Recent methods \cite{gasperini_morbitzer2023md4all, Saunders_2023_ICCV, diffusionadverse, wang2023weatherdepth, diffusion_contrast} target multiple conditions, using data augmentation and generation techniques such as GANs and diffusion models \cite{CycleGAN2017, zheng_2020_forkgan, LDM, controlnet} to create realistic adverse data and employing a teacher-student framework. However, they overlook the importance of learning a common representation and do not consider about bridging the synthetic-real gap. In contrast, we propose an auxiliary space to enhance robust representation learning and introduce a novel real adaptation stage.

% \begin{figure*}[!t]
% \centering
% \includegraphics[width=1.\linewidth, height=.57\linewidth]{figures/figure_pipeline_v2.png}
% % \vspace{-0.1cm}
% \caption{The illustration of our proposed pipeline. }
% \label{fig:pipeline}
% % \vspace{-0.4cm}
% \end{figure*}
\section{Proposed Method}
\label{sec:method}
\begin{figure*}[!t]
\centering
\includegraphics[width=.87\linewidth, height=.38\linewidth]{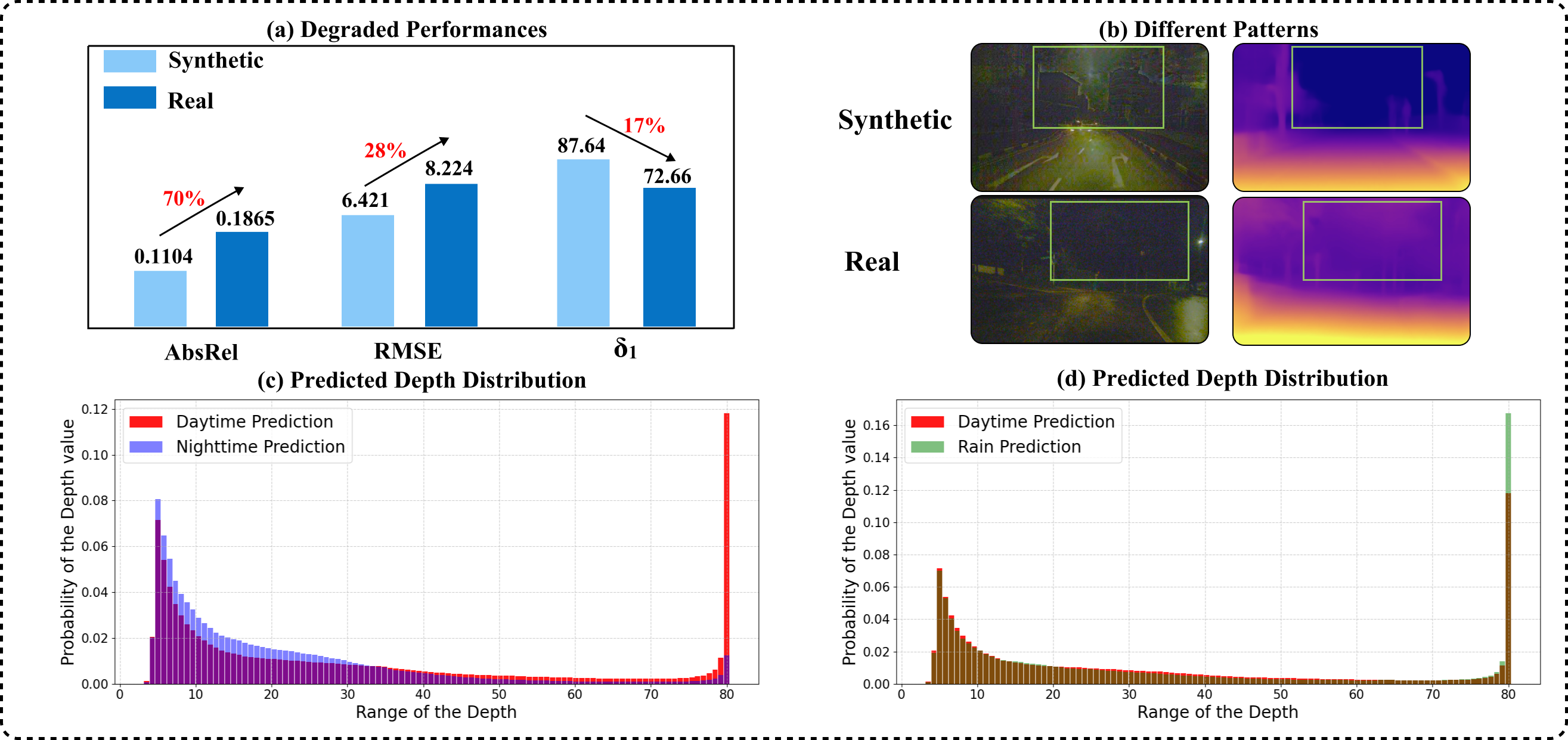}
% \vspace{-0.1cm}
\caption{Analysis on our design. (a) Illustration of the gaps between the synthetic and real data on performance. (b) Green boxes emphasize the different pattern between synthetic and real data, which affects the model's estimation especially for far planes. (c)-(d) The distribution difference between the daytime predicted depth and estimated depth in adverse conditions.}
\label{fig:analysis}
\vspace{-0.2cm}
\end{figure*}

We propose to estimate depth map from single frame or consecutive frames under various conditions, including the normal (e.g., daytime) and adverse scenarios (e.g., rain, nighttime, etc.). We demonstrate our pipeline in Fig. \ref{fig:pipeline}, which consists of a) Synthetic Adaptation (SA); b) Real Adaptation (RA). In SA, we train a network on synthetic data where: 1) daytime images are fed into a pre-trained daytime model to produce the pseudo depth label; 2) clean images are passed by an augmentation process to synthesise adverse images which are used to train a network to estimate depth maps, supervised by pseudo labels. We discuss the technique details in Sec. \ref{method:preliminaries} and Sec. \ref{method:syn}. To further learn the ability to handle real-world adverse images, we design a novel synthetic-to-real process by importing the motion structure priors, pseudo label reweighting strategy as discussed in Sec. \ref{method:real}. Due to limited space, we omit the symbol for the pose network in Fig. \ref{fig:pipeline}.

% Our approach targets to make a single depth model able to work under various kinds of conditions robustly, including the normal and adverse scenarios.
% %
% In definition, given an image that $I_t \in \mathbb{R}^{3 \times w \times h}$ or a consecutive frame pair $I_{t-1}, I_t \in \mathbb{R}^{3 \times w \times h}$, the single depth model can offer an accurate depth map $D_t \in \mathbb{R}^{1 \times w \times h}$, where the images can be sampled from any condition $C_i \in C$, (e.g., $C=\{$daytime, nighttime, rain$\}$). 
% %
% Our pipeline is illustrated in Fig.\ref{fig:pipeline}, where we start from a pretrained multi-frame baseline in daytime in Sec.\ref{method:preliminaries}. 
% %
% After that, the model is adapted to multiple conditions with paired synthetic data and the knowledge transfer though motion and cost volume in Sec.\ref{method:syn}.
% %
% Finally, the model is adapted to multiple real domains with motion and structure priors and a novel pseudo label reweighting strategy in Sec.\ref{method:real}. 
% %
% Due the limited space, we omit the daytime model training and the pose model in Fig.\ref{fig:pipeline}.

\subsection{Preliminaries}
\label{method:preliminaries}
Robust depth estimation is closely connected with self-supervised depth estimation since it can provide a good daytime model for the adaptation. 
For the good performance as well as the motion-structure information in daytime, we choose Manydepth \cite{manydepth} as our daytime baseline and implement on nuScenes and Robotcar datasets \cite{nuscenes2019, robotcar} with all daytime data. 

The pipeline of Manydepth \cite{manydepth} consists of a depth network $\Phi_{D-day}$ and a pose network $\Phi_{P-day}$, the key of which is to inject multi-frame cost volume $cv$ into the model to give explicit geometric guidance. 
The input of the models are consecutive frame pairs $I_{t-1}, I_t$, and the cost volume $cv$ is constructed via $I_{t-1}, I_t$ and the encoder of $\Phi_{D-day}$:
\begin{align}
    cv = \Phi_{D-day}(I_t)&- W(\Phi_{D-day}(I_{t-1}), T, d, K),\label{equ:manydepth_cv}\\
    T  &= \Phi_{P-day}(I_{t}, I_{t-1}),
    \label{equ:manydepth_pose}
\end{align}
where $W(\cdot)$ is the warping operation \cite{xu2023unifying}, $T$ represents the estimated pose, $d$ is the predefined depth candidates and $K$ refers to the camera intrinsics. In short, the depth $D_{t}$ is obtained by
\begin{equation}
     D_{t} = \Phi_{D-day}(I_t, cv).
     \label{equ:manydepth_depth}
\end{equation}

In the training, it also requires a single-frame model as a reference to provide the consistent map $M$ and guidance to the inconsistent parts. The total objective function is
\begin{equation}
     L = (1-M)L_p + L_{consist}+ L_{smooth},
     \label{equ:manydepth_loss}
\end{equation}
where $L_p$ and $L_{smooth}$ are the typical losses for self-supervised depth \cite{godard2019digging, 2017left-right-consistency, 2017cvpr_egomotion}. $L_{consist}$ is a constraint from the single-frame model.
Training on monocular videos provides a strong model to work in normal conditions but suffer from degradation cases, and that's why adapting to both normal and adverse conditions is important.

\subsection{Synthetic Adaptation with Motion and CV}
\label{method:syn}
Since we do not have real daytime-adverse paired data, directly adapting a daytime model to multiple adverse conditions is difficult. 
Following \cite{gasperini_morbitzer2023md4all, Saunders_2023_ICCV, diffusionadverse, wang2023weatherdepth}, we first adapt the pre-trained daytime models $\Phi_{D-day}, \Phi_{P-day}$ (denoted as $\Phi_{day}$ in Fig.\ref{fig:pipeline}) to synthetic adverse model $\Phi_{D-syn}, \Phi_{P-syn}$. 
To realize that, a generative adversarial network (GAN) \cite{zheng_2020_forkgan, CycleGAN2017} or a diffusion model \cite{LDM} defined as $\Phi_{aug}$ is utilized as augmentations to translate the daytime images into multiple adverse conditions as 
\begin{equation}
     I_{C_i-syn} = \Phi_{aug}(I_{day}), C_i \in C,
     \label{equ:manydepth_aug}
\end{equation}
where C represents the set of conditions. Then the adaptation can be conducted on the paired data $I_{day},I_{C_i-syn}$ in a distillation or teacher-student manner, where the outputs of $\Phi_{D-day}, \Phi_{P-day}$  can be defined as $D_{day}, T_{day}$ and those from $\Phi_{D-syn}, \Phi_{P-syn}$ are $D_{syn}, T_{syn}$. 
The distillation objective is the absolute relative depth loss: 
\begin{equation}
     L_{d} = \frac{1}{HW}\sum^{HW}|\frac{D_{day}-D_{syn}}{D_{syn}} |.\label{equ:syn_depthloss}
\end{equation}

However, learning from only depth pseudo labels is still challenging to learn about the adverse patterns \cite{diffusion_contrast, UCDA}, which motivates us to also discover an auxiliary space to inhibit the problems.
%
% It is intuitively to consider about the feature space, but this leads to obvious performance degradation in our experiments. 
% %
% This is due to the abundant semantic and structure information inside feature, and the heterogeneous visual information in different conditions will make it very hard to learn directly from daytime features.
%
Surprisingly, we find that the cost volume can be an auxiliary space to transfer the knowledge.
In theory, the process of constructing cost volume is naturally filtering out the abundant information, which retains explicit motion and structure information and reflects the pixel-level correlation through temporal frames.
This gives us chances to distill explicit and filtered motion and structure knowledge from daytime to adverse conditions.

The cost volumes from $\Phi_{D-day}$ and $\Phi_{D-syn}$ are defined as $cv_{day}, cv_{syn}$. 
Thus, the objective of cost volume learning is added into training: 
\begin{equation}
     L_{cv} = \frac{1}{chw}\sum^{chw}|cv_{day}-cv_{syn}|.\label{equ:syn_cvloss}
\end{equation}

To construct the correct cost volume, we also supervise the $T_{syn}$ by $T_{day}$ with an L2 loss $L_T$. 
The synthetic adaptation is illustrated in Fig.\ref{fig:pipeline} (a), where the total objective $L_{syn}$ is designed with different scale factors:
\begin{equation}
     L_{syn} = \alpha_1 L_d + \alpha_2 L_{cv}+ \alpha_3 L_{T}.
     \label{equ:syn_totalloss}
\end{equation}

We also notice that the cost volume constructed by feature dot product can lead to more robust learning compared with feature difference.
Thus Eq.\ref{equ:manydepth_cv} will be changed into $cv = \Phi_{D-day}(I_t) \cdot W(\Phi_{D-day}(I_{t-1}), T, d, K)$.

% \begin{figure*}[!t]
% \centering
% \includegraphics[width=.9\linewidth]{figures/figure_analysis.png}
% % \vspace{-0.1cm}
% \caption{Illustration of the gaps between the synthetic and real data. }
% \label{fig:analysis}
% % \vspace{-0.4cm}
% \end{figure*}

\subsection{Real Adaptation with Consistency Reweighting and Structure Prior}
\label{method:real}

% Although the synthetic adaptation importantly bridges the gap between daytime and multiple synthetic adverse conditions, there are inevitable gaps between synthetic adverse and real adverse conditions \cite{gasperini_morbitzer2023md4all, Saunders_2023_ICCV, UCDA}. 
% %
% Here we analyze the performance of the models $\Phi_{D-syn}, \Phi_{P-syn}$ on synthetic data and real data of the same condition (e.g., night), and visualize it in Fig.\ref{fig:analysis} (a) and (b).
% %
% Notably, (a) shows that the performance degrades significantly from the synthetic adverse data to real adverse data in all metrics (degraded by $70\%$ in $AbsRel$, $28 \%$ in $RMSE$ and $17\%$ in $\delta_1$).
% %
% As displayed in (b), the synthetic nighttime image with its predicted depth map, is in different patterns compared with the real nighttime, such as the model is able to capture the near and far distances of a synthetic image, but can not detect the far planes in the real one.
% %
% Inspired by this, our next mission is to solve the problem with the synthetic-real domain gap. 

Although synthetic adaptation helps to bridge the gap between daytime and synthetic adverse conditions, there are still inevitable gaps between synthetic and real adverse conditions \cite{gasperini_morbitzer2023md4all, Saunders_2023_ICCV, UCDA}. Here, we analyze the performance of models $\Phi_{D-syn}$ and $\Phi_{P-syn}$ on synthetic and real data under the same condition (e.g., night), as shown in Fig.\ref{fig:analysis} (a) and (b).
In (a), performance significantly declines from synthetic to real adverse data, with drops of $70\%$ in $AbsRel$, $28\%$ in $RMSE$, and $17\%$ in $\delta_1$. In (b), the synthetic nighttime image and its predicted depth map show different patterns compared to the real nighttime, such as successful detection of near and far distances in synthetic images, and failure to detect distant planes in real images.
This motivates us to address the synthetic-real domain gap.

% Formally, our target is to train a depth model $\Phi_{D-real}$ and a pose model $\Phi_{P-real}$ that can predict robust $D_{real}$ in diverse kinds of real conditions without requiring any labels.
% %
% To realize that, we simultaneously utilize the ability of $\Phi_{D-day}$ and $\Phi_{D-syn}$, to elaborately design a \textbf{Real Adaptation} stage, as illustrated in Fig.\ref{fig:pipeline} (b).

Formally, our goal is to train a depth model $\Phi_{D-real}$ and a pose model $\Phi_{P-real}$ capable of predicting $D_{real}$ across various real conditions without any labels.
To achieve this, we leverage the abilities of $\Phi_{D-day}$ and $\Phi_{D-syn}$ by carefully designing a Real Adaptation stage in Fig. \ref{fig:analysis} (b).

% \paragraph{Consistency Reweighting Strategy}
\vspace{0.2cm}
\noindent \textbf{Consistency Reweighting Strategy}
% A direct supervision signal for $D_{real}$ is $D_{syn}$ which is from the $\Phi_{D-syn}$ on real data and can be simply utilized as the pseudo label for $\Phi_{D-real}$. 
% %
% However, since $\Phi_{D-syn}$ fits to the synthetic domain well and can not perfectly work in real adverse conditions, an intuitive way is to emphasize the supervision on the trustworthy parts. 
% %
% Considering this, the $\Phi_{D-day}$ is also leveraged to provide predictions $D_{day}$ on the real images of diverse conditions.
% %
% Our strategy is to compare $D_{day}$ and $D_{syn}$ and then give larger weights to the consistent parts while set smaller weights to the very inconsistent regions.
% %
% With the exponential function \cite{prodepth}, we can effectively design a confidence map $C_{consistency}$ based on the consistency, and transform the confidence map to reweighting map $W_{consistency}$:
A straightforward supervision signal for $D_{real}$ is $D_{syn}$, generated by $\Phi_{D-syn}$ on real data, which can serve as a pseudo label for $\Phi_{D-real}$. However, since $\Phi_{D-syn}$ is optimized for the synthetic domain and may not perform well in real adverse conditions, it is intuitive to emphasize supervision on the more reliable parts.
To address this, we also leverage $\Phi_{D-day}$ to provide predictions $D_{day}$ on real images across diverse conditions. Our strategy compares $D_{day}$ and $D_{syn}$, assigning higher weights to consistent regions and lower weights to highly inconsistent areas.
Using an exponential function \cite{prodepth}, we design a confidence map $C_{cst}$ based on the consistency and transform it into a reweighting map $W_{cst}$:
\begin{align}
    C_{cst} &= e^{-\beta\frac{|D_{syn}-D_{day}|}{D_{syn}}},
    \label{equ:consistency}\\
    W_{cst} &= C_{cst}+ \epsilon.
    \label{equ:consistency_reweighting}
\end{align}

The $\beta$ is a scale factor and the $\epsilon$ is a weight bias to guarantee a weak supervision. 
The supervision for $D_{real}$ will be changed into a reweighting consistent-depth loss $L_{cd}$:
\begin{equation}
     L_{cd} = \frac{1}{HW}\sum^{HW}W_{cst}|\frac{D_{real}-D_{syn}}{D_{real}} | .
     \label{equ:consistent-depth-loss}
\end{equation}

\noindent \textbf{Structure Prior Constraint}
% Although learning from pseudo-label is reweighted, it still can not perfectly supervise the real adverse predictions, which motivates us to seek for more explicit priors from daytime data and depth. 
% %
% Digging into this, the distributions of the predicted depths from $\Phi_{D-day}$ and $\Phi_{D-syn}$ on the real training set are calculated and visualized in Fig.\ref{fig:analysis} (c) and (d).
% %
% In (c), we surprisingly find that the dominant distribution difference between nighttime and daytime is depths at the very far planes (e.g., near $80$ which is the upper limit of model prediction).
% %
% It is observed that more pixels of nighttime depth will be predicted to the near planes ($<30$) compared with daytime depth. 
% %
% This conforms with the bad visibility of objects at far distance in nighttime, which will lead the model to judge more regions as the near ones.
% %
% Differently, the obvious difference between daytime and rain distributions in Fig.\ref{fig:analysis} (d) is the model predicts more regions with large depth. This is due the severe blurry parts of rain condition which will mislead the model to predict the near objects as far away.
Although pseudo-label learning is reweighted, it still cannot perfectly supervise real adverse predictions, motivating us to seek more explicit priors from daytime data and depth.
To investigate further, we calculate and visualize the predicted depth distributions of $\Phi_{D-day}$ and $\Phi_{D-syn}$ on the real training set in Fig.\ref{fig:analysis} (c) and (d). In (c), we observe that the primary distribution difference between nighttime and daytime is at the far planes (e.g., near $80$, the model’s upper prediction limit). Notably, nighttime depth contains more near-plane predictions ($<30$) compared to daytime depth, consistent with poor visibility at long distances under nighttime conditions, causing the model to predict more areas as closer.
In contrast, Fig.\ref{fig:analysis} (d) shows that under rainy condition, the model predicts more distant regions, likely due to rain-induced blur misleading the model to perceive near objects as farther away.

% Inspired by all these, we set the daytime predicted distribution as a reference distribution and target to align all distributions of adverse conditions to that of daytime. 
% %
% Formally, the distribution of daytime and adverse conditions can be defined as $Dist_{day}$, $Dist_{adverse}$, which are obtained from the predicted values through depth bins.
% %
% However, directly calculating the distributions will make the histogram indifferentiable.
% %
% Inspired by \cite{differentiable_hist, 2019huenet}, we utilize the differentiable histograms to caputre the distributions.
% %
% Setting $N$ sub-intervals between $d_{min}$ and $d_{max}$, where each interval has length of $L=\frac{d_{max}-d_{min}}{N}$ and the center for interval $n$ is $b_n=d_{min}+L(n+\frac{1}{2})$. 
% %
% By utilizing two sigmoid functions $\sigma$ to model the kernel function $K(x)=\sigma(x)\sigma(-x)$, the probability of a depth map $D$ in interval $k$ is
Inspired by these insights, we set the daytime predicted distribution as a reference and aim to align the distributions of adverse conditions with that of daytime.
Formally, the distributions for daytime and adverse conditions are defined as $Dist_{day}$ and $Dist_{adv}$, obtained from the predicted depth values across defined depth bins. However, direct calculation of these distributions makes the histogram indifferentiable.
Following \cite{differentiable_hist, 2019huenet}, we use differentiable histograms to capture these distributions. We define $N$ sub-intervals between $d_{min}$ and $d_{max}$, where each interval has a length of $L=\frac{d_{max}-d_{min}}{N}$, with center $b_n=d_{min}+L(n+\frac{1}{2})$ for interval $n$.
Using two sigmoid functions $\sigma$ to model the kernel function $K(x)=\sigma(x)\sigma(-x)$, the probability of a depth map $D$ falling within interval $k$ is
\begin{align}
     P(n) =\frac{1}{HW}\sum_{x=1}^{HW}[&\sigma(\frac{D(x)-b_n+\frac{L}{2}}{a}) \notag\\
     &-\sigma(\frac{D(x)-b_n-\frac{L}{2}}{a})],
     \label{equ:differential_histogram}
\end{align}
where $P$ is the differentiable probability function and $a$ is a bandwidth factor. Finally, the $Dist_{day}$, $Dist_{adv}$ are consist of $P_{day}, P_{adv}$:
\begin{align}
     Dist_{day} &=  \{b_n, P_{day}(n)\}_{n=1}^{N},\label{equ:distribution_day}\\
     \quad Dist_{adv} &=  \{b_n, P_{adv}(n)\}_{n=1}^{N}.
     \label{equ:distribution_adverse}
\end{align}

Noted that $Dist_{day}$ is calculated among the whole daytime training set instead of one sample. The objective is to minimize the distribution distances between $Dist_{day}$ and $Dist_{adv}$ by K-L divergence:
\begin{equation}
     L_{dis}^{KL} = \sum_{i=1}^{N}P_{adv}(i) \cdot log \frac{P_{adv}(i)}{P_{day}(i)}.
     \label{equ:kl_loss}
\end{equation}

Apart from $L_{cd}$ and $L_{dis}^{KL}$, we also utilize $L_{cv}$ and $L_{T}$ as in previous stage. Thus, the total objective function $L_{real}$ is 
\begin{equation}
     L_{real} = \alpha_1 L_{cd} + \alpha_2 L_{dis}^{KL} + \alpha_3 L_{cv}
     + \alpha_4 L_{T}
     \label{equ:real_loss}
\end{equation}

With the designed learning procedure, our model can gradually fit better to multiple real conditions, rather than only learn from the synthetic conditions. 
\section{Experiments}
\label{sec:experiments}

\begin{table*}[!t] 
\renewcommand{\arraystretch}{1.2}

\centering
% \Huge
\footnotesize
% \small
\setlength{\tabcolsep}{4pt}  % 调整列间距
\begin{tabular}{l|c|cccc|cccc|cccc} %需要10列
\toprule %添加表格头部粗线
\cline{1-14}\
\multirow{2}{*}{Method} & \multicolumn{1}{c|}{Test} & \multicolumn{4}{c|}{nuScenes-day} & \multicolumn{4}{c|}{nuScenes-night} & \multicolumn{4}{c}{nuScenes-rain}\\
     &frames& AbsRel$\downarrow$ & SqRel$\downarrow$ &  RMSE$\downarrow$ & $ \delta_1\uparrow $ & AbsRel$\downarrow$ & SqRel$\downarrow$ &  RMSE$\downarrow$ & $ \delta_1\uparrow $ & AbsRel$\downarrow$ & SqRel$\downarrow$ &  RMSE$\downarrow$ & $ \delta_1\uparrow $    \\  
% \midrule
\cline{1-14}
Monodepth2 \cite{godard2019digging} & 1 &0.1333 &1.820 &6.459 &\underline{85.88} &0.2419 &2.776 &10.922 &58.17 & 0.1572 &2.273 &7.453 & 79.49 \\
R4Dyn$^{\textbf{\dag}}$ (\textbf{Radar}) \cite{gasperini2021r4dyn} & 1 &\underline{0.1259} &\underline{1.661} &\underline{6.434} &\textbf{86.97} &0.2194 & 2.889  &10.542 &62.28 &\underline{0.1337} &1.938 &7.131 &\textbf{83.91} \\
RNW \cite{wang2021RNW} & 1 &0.2872 &3.433  &9.185 &56.21 &0.3333 &4.066 &10.098 &43.72 &0.2952 &3.796 &9.341 &57.21 \\
 md4all-AD \cite{gasperini_morbitzer2023md4all}   & 1 &0.1523 &2.141 &6.853 &83.11 &0.2187 &2.991 &9.003 &68.84 &0.1601 &2.259 &7.832 &78.97 \\
 robust-depth \cite{Saunders_2023_ICCV}  & 1 &0.1436 &1.862 &6.802 &82.95 &0.2101 &2.691 &8.673 &69.58 &0.1458 &1.891 &7.371 &80.21 \\
 md4all-DD \cite{gasperini_morbitzer2023md4all} & 1 &0.1366 &1.752 &6.452 &84.61 &0.1921 & \underline{2.386} &8.507 &71.07 &0.1414 &\underline{1.829} &7.228 &80.98 \\
 DM-MDE \cite{diffusionadverse} & 1 &0.1280 &- &6.449 &84.03 &\underline{0.1910} &- &\underline{8.433} &\textbf{71.14} &0.1390 &- &\underline{7.129} &\underline{81.36} \\
 \textbf{Ours-Single} & 1 & \textbf{0.1239} &\textbf{1.454} &\textbf{6.214} &85.66 &\textbf{0.1792} &\textbf{1.896} &\textbf{7.949} &\underline{71.08} &\textbf{0.1331} &\textbf{1.616} &\textbf{6.926} &\textbf{81.68} \\
% \midrule
\cline{1-14}
 Manydepth \cite{manydepth} & 2(-1,0) &0.1213 &1.699 &6.183 &\textbf{87.05} &0.2442 &3.347 &11.38 &57.04 &0.1347 &1.730 &7.110 &81.24 \\
 Manydepth \cite{manydepth}+\cite{gasperini_morbitzer2023md4all} & 2(-1,0) &0.1202 &1.417 &6.111 &86.62 &0.1865 &2.255 &8.245 &72.58&0.1310 &1.717 &6.840 &82.61 \\
\textbf{Ours-Multi} & 2(-1,0) &\textbf{0.1170} &\textbf{1.371} &\textbf{6.023} &86.66 & \textbf{0.1713} & \textbf{1.779} &\textbf{7.689} &\textbf{73.16} &\textbf{0.1266} &\textbf{1.532} &\textbf{6.654} &\textbf{82.86} \\
\cline{1-14}
\bottomrule 
\end{tabular}
\caption{Quantitative comparison on nuScenes in different conditions (daytime, nighttime and rain). $\textbf{\dag}$ indicates the method \cite{gasperini2021r4dyn} using \textbf{additional Radar information}. Test frames indicates the evaluation is conducted in multi-frame or single-frame setting. - points to the missing value from the reference. The first and second ranked performances are indicated by \textbf{bold} and \underline{underline}.}
\label{tab:nuscenes}
\vspace{-0.2cm}
\end{table*}

\begin{table*}[htb] 
\renewcommand{\arraystretch}{1.2}
\centering
\footnotesize
% \small
% \Large
% \setlength{\tabcolsep}{1pt}  % 调整列间距
\resizebox{.85\linewidth}{!}{
\begin{tabular}{l|c|c|cccc|cccc} %需要10列
\toprule %添加表格头部粗线
\cline{1-11}
\multirow{2}{*}{Method } & Source & \multicolumn{1}{c|}{Test} & \multicolumn{4}{c|}{Robotcar-day} & \multicolumn{4}{c}{Robotcar-night}\\

                        &  & frames & AbsRel$\downarrow$ & SqRel$\downarrow$ &  RMSE$\downarrow$ & $ \delta1\uparrow $ & AbsRel$\downarrow$ & SqRel$\downarrow$ &  RMSE$\downarrow$ & $ \delta1\uparrow $  \\  
\cline{1-11}
   Monodepth2 \cite{godard2019digging}                     &\cite{gasperini_morbitzer2023md4all} &1 &0.1209 &0.723 &3.335 &86.61 &0.3909 &3.547 &8.227 &22.51 \\
  DeFeatNet \cite{spencer2020defeat}     &\cite{vankadari2023sun}   &1 &0.2470 &2.980 &7.884 &65.00 &0.3340 &4.589  &8.606 &58.60 \\
  ADIDS \cite{liu2021ADIDS}      &\cite{vankadari2023sun} &1&0.2390 &2.089 &6.743 &61.40 &0.2870 &2.569  &7.985 &49.00 \\
   RNW \cite{wang2021RNW}                 &\cite{vankadari2023sun}    &1 &0.2970 &2.608  &7.996 &43.10 &0.1850 &1.710 &6.549 &73.30  \\
   WSGD  \cite{vankadari2023sun}                &\cite{vankadari2023sun}       &1&0.1760 &1.603 &6.036 &75.00 &0.1740 &1.637 &6.302 &75.40 \\
   robust-depth \cite{Saunders_2023_ICCV}             &[self]        & 1 &0.1225 &0.776 &3.302 &85.84 &0.1333 &0.895 &3.756 &85.11  \\
   md4all \cite{gasperini_morbitzer2023md4all}     &\cite{gasperini_morbitzer2023md4all}    & 1 &0.1128 &\underline{0.648} & \underline{3.206} &87.13 &\underline{0.1219} &0.784 &\underline{3.604} &84.86 \\
   DM-MDE \cite{diffusionadverse}      &\cite{diffusionadverse} & 1 &0.1190 &0.728 &3.287 &87.17 &0.1290 &\underline{0.751} &3.661 &83.86 \\
   % \textbf{Ours}                     &\textbf{0.1237} &\textbf{1.349} &\textbf{6.317} &84.77 &\textbf{0.1841} &\textbf{2.087} &\textbf{8.275} &\underline{70.75} &\textbf{0.1332} &\textbf{1.613} &\textbf{7.022} &80.86 \\
   \cline{1-11}
   Manydepth \cite{manydepth}     &[self]  & 1 &0.1200 &0.802 &3.413 &87.46 &0.4974 &5.517 &10.21 &14.87 \\
   Manydepth \cite{manydepth}+\cite{gasperini_morbitzer2023md4all}    &[self]   & 1 &\underline{0.1092} &0.759 &3.355 &\textbf{88.32} &0.1257 &0.930 &3.904 &\underline{85.24} \\
   \textbf{Ours-Single}                &[self]     & 1 &\textbf{0.1063} &\textbf{0.646} &\textbf{3.171} &\underline{88.29} &\textbf{0.1103} &\textbf{0.693} &\textbf{3.455} &\textbf{86.15}  \\
\hline
\bottomrule 
\end{tabular}}
\caption{Quantitative comparison on Robotcar dataset \cite{robotcar} in different conditions (daytime and nighttime). Source indicates where the metrics are cited, and [self] refers to our own implementations.}
\label{tab:robotcar}
\vspace{-0.4cm}
\end{table*}

\begin{table}[htb] 
\renewcommand{\arraystretch}{1.1}  % 增加行距
\setlength{\tabcolsep}{3pt}  % 增加列间距

\centering
\footnotesize
\begin{tabular}{l|c|cccc}
\toprule
\multirow{2}{*}{ Method} &Test & \multicolumn{4}{c}{DrivingStereo-rain} \\
\cline{3-6}
&frames& AbsRel$\downarrow$ & SqRel$\downarrow$ & RMSE$\downarrow$ & $\delta_1\uparrow$ \\
\midrule
robust-depth \cite{Saunders_2023_ICCV} &1& 0.2622 & 4.309 & 11.657 & 59.58 \\
md4all \cite{gasperini_morbitzer2023md4all} &1& 0.1822 & 2.007 & 8.465 & 70.35 \\
Manydepth \cite{manydepth}+\cite{gasperini_morbitzer2023md4all} &1& 0.1852 & 2.020 & 8.515 & 69.24 \\
\textbf{Ours-Single} &1& \textbf{0.1707} & \textbf{1.755} & \textbf{8.004} & \textbf{72.98} \\
\toprule
\multirow{2}{*}{ Method} &Test & \multicolumn{4}{c}{DrivingStereo-fog} \\
\cline{3-6}
&frames& AbsRel$\downarrow$ & SqRel$\downarrow$ & RMSE$\downarrow$ & $\delta_1\uparrow$ \\
\midrule
robust-depth \cite{Saunders_2023_ICCV} &1& 0.1366  & 1.260  & 6.876 & 84.01 \\
md4all \cite{gasperini_morbitzer2023md4all} &1& \textbf{0.1259} & 1.081 & \textbf{6.383} & \textbf{86.16} \\
Manydepth \cite{manydepth}+\cite{gasperini_morbitzer2023md4all} &1& 0.1335 & 1.162 & 6.740 & 83.50 \\
\textbf{Ours-Single} &1& 0.1281 & \textbf{1.023} & 6.422 & 84.52 \\
\bottomrule
\end{tabular}
\caption{Zero-shot evaluation on DrivingStereo (rain, fog). Noted that our method is designed for both single-frame and multi-frame settings, and definitely defeats the compared baseline \cite{manydepth}+\cite{gasperini_morbitzer2023md4all}.}
\label{tab:drivingstereo}
\vspace{-0.6cm}
\end{table}

\begin{figure*}[!t]
    \centering
    % 第一部分: nuScenes
    \begin{tabular}{c @{} c @{} c @{} c @{} c @{} c @{} c }
        % 第一行标题: UDA
        \multirow{2}{*}[3mm]{\rotatebox[origin=c]{90}{\textbf{nuScenes}}} \hspace{1pt}&
        % \multirow{2}{*}{\rotatebox[origin=c]{90}{\textbf{nuScenes}}} \hspace{1pt}&
        \multirow{1}{*}[9mm]{\rotatebox[origin=c]{90}{\textbf{night}}} \hspace{1pt}&
        \includegraphics[width=0.18\textwidth]{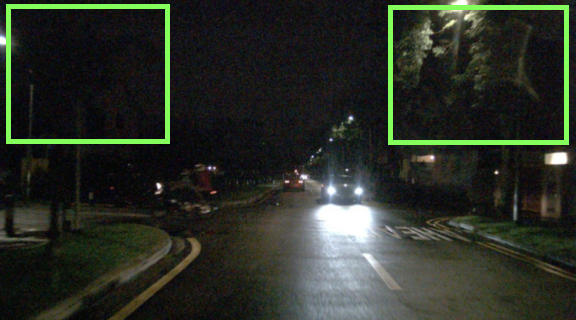} \hspace{1pt} &
        \includegraphics[width=0.18\textwidth]{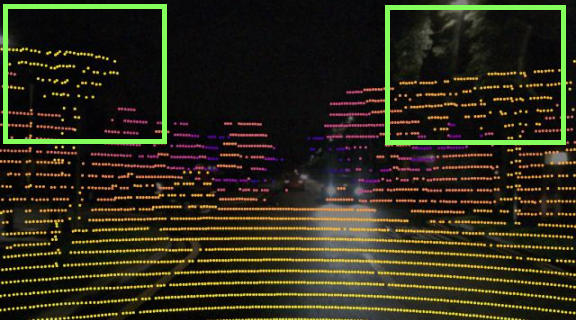} \hspace{1pt} &
        \includegraphics[width=0.18\textwidth]{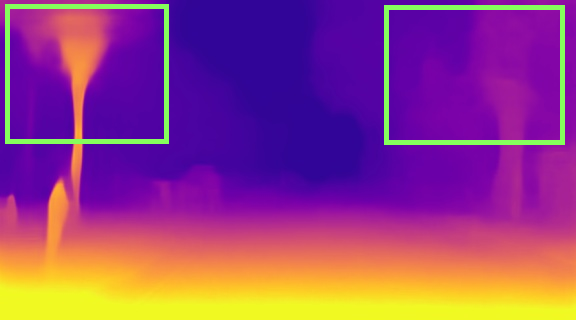} \hspace{1pt} &
        \includegraphics[width=0.18\textwidth]{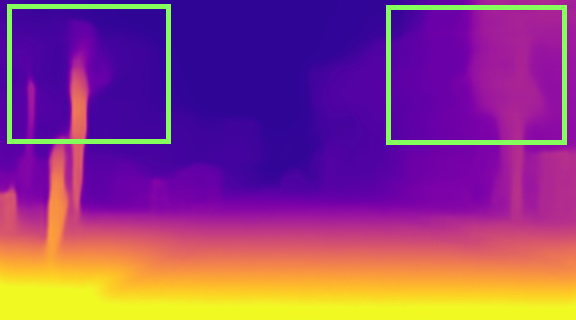} \hspace{1pt} &
        % \includegraphics[width=0.15\textwidth]{figures/figures_nuscenes_drivingstereo/nuscenes_night_manydepth.png} 
        % \hspace{1.5pt} &
        \includegraphics[width=0.18\textwidth]{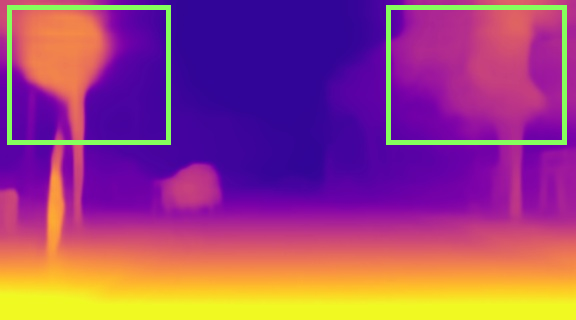} \\
        
        % % 第二行: Rain
        & \multirow{1}{*}[9mm]{\rotatebox[origin=c]{90}{\textbf{rain}}} \hspace{1pt}&
        \includegraphics[width=0.18\textwidth]{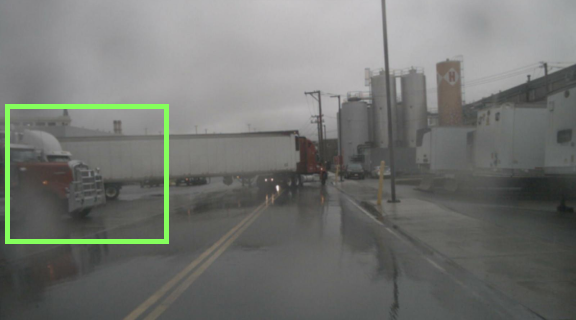} \hspace{1pt}&
        \includegraphics[width=0.18\textwidth]{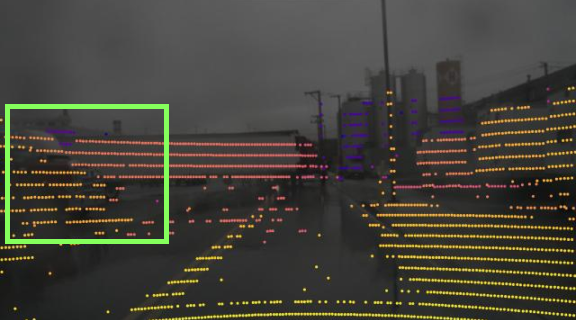} \hspace{1pt}&
        \includegraphics[width=0.18\textwidth]{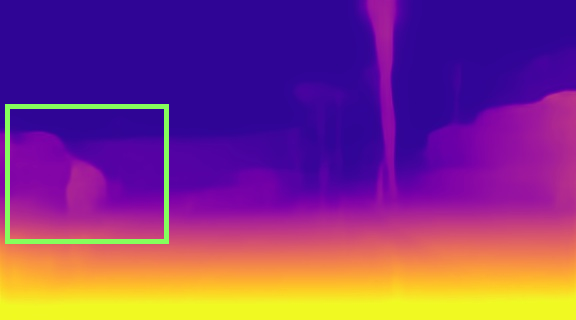} \hspace{1pt}&
        \includegraphics[width=0.18\textwidth]{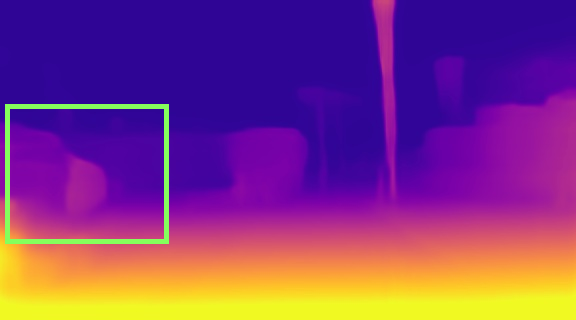} \hspace{1pt}&
        \includegraphics[width=0.18\textwidth]{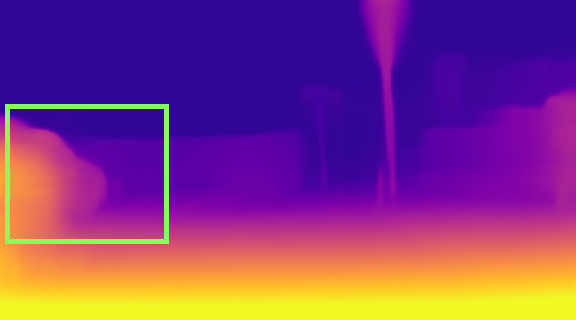} \\
        
        % % 第二部分: DrivingStereo
        \multirow{2}{*}[8mm]{\rotatebox[origin=c]{90}{\textbf{DrivingStereo}}} \hspace{1pt}&
        \multirow{1}{*}[9mm]{\rotatebox[origin=c]{90}{\textbf{fog}}} \hspace{1pt}&
        \includegraphics[width=0.18\textwidth]{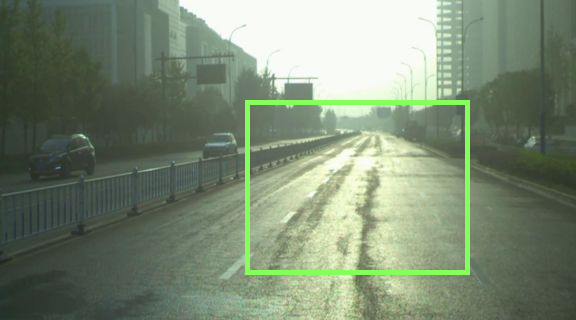} \hspace{1pt} &
        \includegraphics[width=0.18\textwidth]{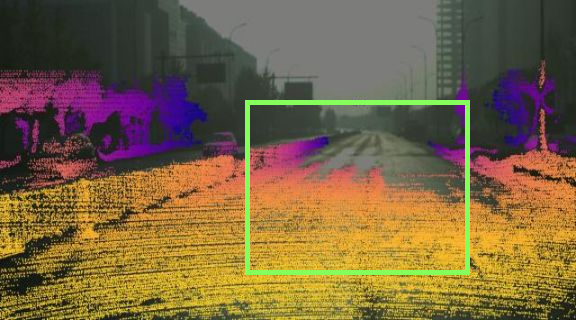} \hspace{1pt} &
        \includegraphics[width=0.18\textwidth]{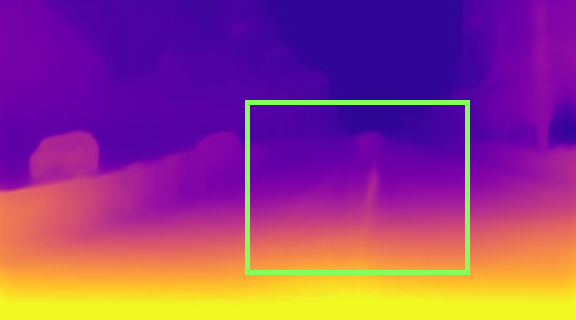} \hspace{1pt} &
        \includegraphics[width=0.18\textwidth]{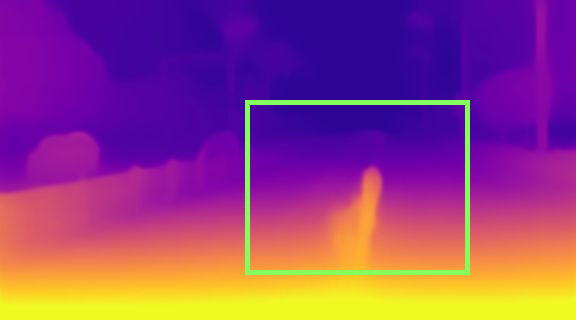} \hspace{1pt} &
        % \includegraphics[width=0.15\textwidth]{figures/figures_nuscenes_drivingstereo/nuscenes_night_manydepth.jpg} 
        % \hspace{1.5pt} &
        \includegraphics[width=0.18\textwidth]{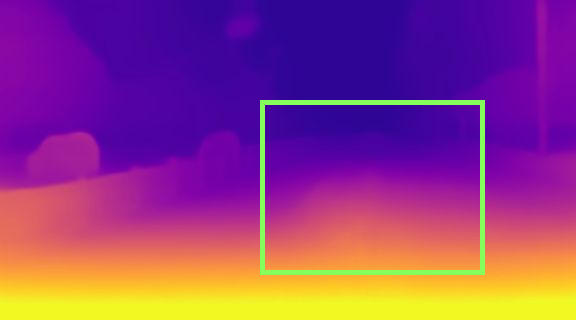} \\
        
        % % 第二行: Rain
        & \multirow{1}{*}[9mm]{\rotatebox[origin=c]{90}{\textbf{rain}}} \hspace{1pt}&
        \includegraphics[width=0.18\textwidth]{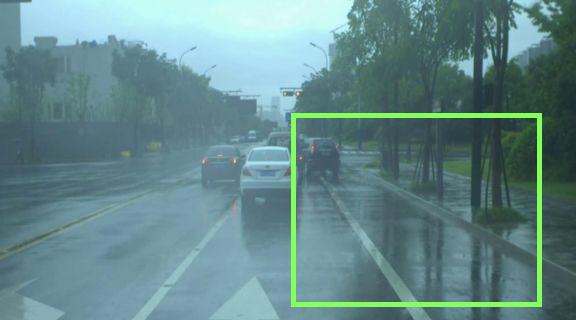} \hspace{1pt}&
        \includegraphics[width=0.18\textwidth]{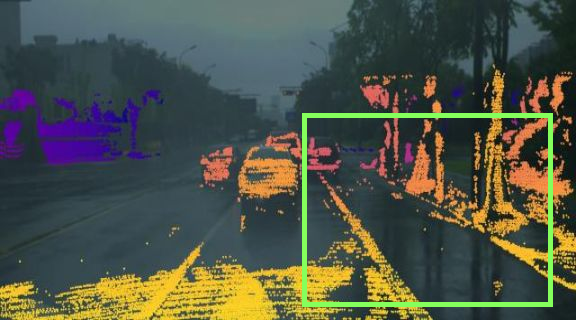} \hspace{1pt}&
        \includegraphics[width=0.18\textwidth]{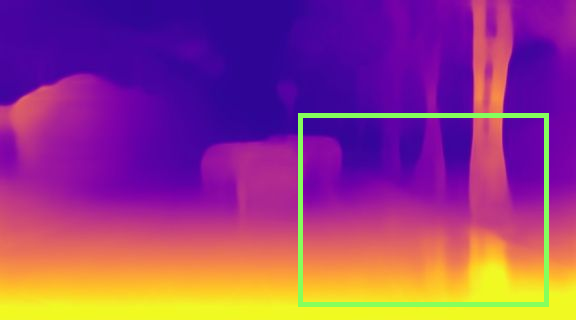} \hspace{1pt}&
        \includegraphics[width=0.18\textwidth]{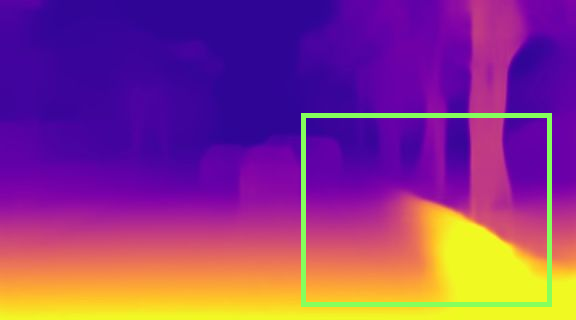} \hspace{1pt}&
        \includegraphics[width=0.18\textwidth]{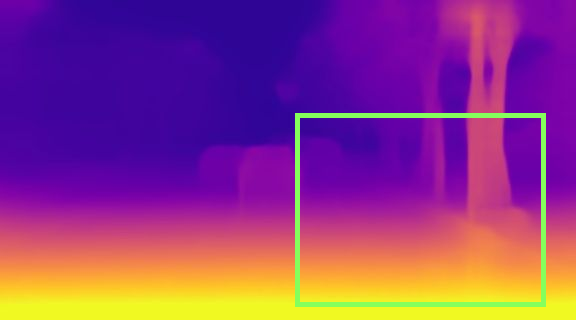} \\
        
        % 底部的标签行
         &&Input Image & Ground Truth & robust-depth \cite{Saunders_2023_ICCV} & md4all \cite{gasperini_morbitzer2023md4all}  & Ours \\
    \end{tabular}

    % 添加总体标题
    \caption{Qualitative comparison of depth predictions under adverse conditions. The first two rows refer to nighttime and rain conditions in nuScenes \cite{nuscenes2019} dataset, and the last two rows are zero-shot evaluation results on DrivingStereo \cite{drivingstereo} dataset. Our method addresses the challenging cases indicated in green boxes (can be checked via the ground truth), where other methods fail.} 
    \label{fig:qualitative_nuscenes_drivingstereo}
    \vspace{-0.2cm}
\end{figure*}

\subsection{Implementation Details}
\label{sec:implement}
The experiments on all models of different stages are conducted on the same ResNet architecture \cite{resnet}, so do the baselines. The training for the $\Phi_{day}$ and the $\Phi_{syn}$ are conducted for 20 epochs, while $\Phi_{real}$ is initialized by $\Phi_{syn}$ and trained for 10 epochs. The learning rate is initialized to $1\times 10^{-4}$ with a decreasing factor of 0.5 every 5 epochs.  For the adaptive depth bins in Manydepth \cite{manydepth}, we fix them to $3.5-80m$ since \cite{gasperini_morbitzer2023md4all, diffusionadverse, packnet-sfm} all use a weak velocity loss to align the scales. Thanks to \cite{gasperini_morbitzer2023md4all, diffusionadverse}, we can utilize the synthetic data from their pretrained augmentation models. For the $Dist_{day}$, we obtain a fixed distribution through the whole daytime training dataset, and utilize it for the real adaptation constraint. Similar to \cite{gasperini_morbitzer2023md4all, Saunders_2023_ICCV}, our training is conducted with a mixture of daytime and adverse data, and more details can be found in the supplementary.

% The training of all the pipeline is conducted on a single NVIDIA A5000 GPU.

\subsection{Datasets}
\label{sec:datasets}
Our approach is evaluated on three mainstream public datasets for robust depth estimation, including nuScenes \cite{nuscenes2019} and Robotcar \cite{robotcar} which are used for training and testing, and DrivingStereo \cite{drivingstereo} dataset which is used for zero-shot evaluation to verify the effectiveness of our pipeline.

\noindent \textbf{NuScenes} \cite{nuscenes2019} is a large benchmark which owns real data in daytime, nighttime, and rain conditions with 15 hours of driving. Following \cite{gasperini_morbitzer2023md4all, diffusionadverse, gasperini2021r4dyn}, a total of 21476 images (including 15129 in daytime, 3796 in rain, 2551 in nighttime) are considered for training. For evaluation, the set owns 6019 consecutive frame pairs (4449 in daytime, 1088 in rain, 602 in nighttime). Our model is evaluated both under single-frame and multi-frames settings. All the metrics are computed for ground truth within $0.1-80m$.

\noindent \textbf{Robotcar} \cite{robotcar} contains daytime and nighttime driving data. We follow the split of WSGD \cite{vankadari2023sun}, where there are 17790 images in daytime and 19612 in nighttime. The test set includes 1411 single images, in which 702 are daytime and 709 are nighttime. Our model is evaluated in single-frame setting on the dataset, with the depth range of $0.1-50m$.

\noindent \textbf{DrivingStereo} \cite{drivingstereo} is a dataset that includes the real rainy and foggy images, where each condition includes 500 single images for our zero-shot evaluation. We follow the use of data in \cite{wang2023weatherdepth} to evaluate our model in single-frame setting and all metrics are computed within a range of $0.1-80m$.

\subsection{Evaluation Metrics and Baselines}
\label{sec:metrics_and_baselines}
Following previous methods \cite{godard2019digging,manydepth,gasperini_morbitzer2023md4all, diffusionadverse}, we set the absolute relative error (AbsRel), square relative error (SqRel), root mean square error (RMSE), and threshold accuracy ($\delta_1$) as the evaluation protocol.
We choose the competitive state-of-the-art methods as our baselines, including \cite{gasperini_morbitzer2023md4all, diffusionadverse, Saunders_2023_ICCV, wang2021RNW, vankadari2023sun} with their baseline \cite{godard2019digging}, the Radar-based method \cite{gasperini2021r4dyn}, and \cite{liu2021ADIDS, spencer2020defeat}. Since our multi-frame training pipeline is built on \cite{manydepth}, we also compare with \cite{manydepth} and a self-designed baseline \cite{manydepth}+\cite{gasperini_morbitzer2023md4all} (using only synthetic adverse data with pseudo labels to train Manydepth).

\subsection{Quantitative Experimental Results}
\label{sec:quantitative_exp}

\begin{table*}[htb] 
\renewcommand{\arraystretch}{1.2}
\centering
\footnotesize
% \small
% \Large
% \setlength{\tabcolsep}{1pt}  % 调整列间距
\resizebox{.92\linewidth}{!}{
\begin{tabular}{cccccc|cccc|cccc} %需要10列
\toprule %添加表格头部粗线
\cline{1-14}
\multicolumn{6}{c|}{Modules }  & \multicolumn{4}{c|}{nuScenes-night} & \multicolumn{4}{c}{nuScenes-rain}\\

                         \textbf{SA} & $L_{feat}$& $L_{cv}$ & \textbf{RA} & $\textbf{CR}$ & $L_{dis}^{KL}$ & AbsRel$\downarrow$ & SqRel$\downarrow$ &  RMSE$\downarrow$ & $ \delta1\uparrow $ & AbsRel$\downarrow$ & SqRel$\downarrow$ &  RMSE$\downarrow$ & $ \delta1\uparrow $  \\  
\cline{1-14}
    \checkmark &&&&&& 0.1865 &2.255 &8.245 &72.58&0.1310 &1.717 &6.840 &82.61         \\
    \checkmark&\checkmark&&&&& 0.1911& 2.495& 8.456& 72.41& 0.1335 & 1.756 & 7.056 & 82.04 \\
    \checkmark&&\checkmark&&&& 0.1809& 2.019& 7.988& 72.61& 0.1290 & 1.657&6.794 & 82.76 \\
           \checkmark&&\checkmark&\checkmark& && 0.1781 & 1.892 & 7.896 & 72.45 & 0.1288 & 1.610 & 6.856 & 82.36     \\
        \checkmark&&\checkmark&\checkmark&\checkmark&& 0.1743 &1.809 & 7.716 & 73.10 &0.1256  &1.576 &6.743& 82.82       \\
      \checkmark&&\checkmark&\checkmark&\checkmark&\checkmark&  \textbf{0.1713} & \textbf{1.779} &\textbf{7.689} &\textbf{73.16} &\textbf{0.1266} &\textbf{1.532} &\textbf{6.654} &\textbf{82.86}          \\
  
\hline
\bottomrule 
\end{tabular}}
\vspace{-0.1cm}
\caption{Ablations on each module and stage mentioned in the pipeline. The checkmark represents utility of the corresponding module. SA refers to the synthetic adaptation with only depth supervision, and RA is training on real data with only pseudo label in real adaptation.}
\label{table:ablation}
\vspace{-0.3cm}
\end{table*}

Our quantitative results across different datasets are shown in Tables \ref {tab:nuscenes}-\ref{tab:drivingstereo}. Tab. \ref {tab:nuscenes} presents evaluations on the nuScenes dataset \cite{nuscenes2019} under various conditions (daytime, nighttime, rain) in single-frame (two inputs of the model are the same frame) and multi-frame settings (two inputs are consecutive frames), where our method significantly outperforms prior approaches. Specifically, in the single-frame setting, we achieve average improvements of $4.8\%, 16.8\%, 4.2\%$ on AbsRel, SqRel, and RMSE, respectively, over recent methods \cite{gasperini_morbitzer2023md4all, diffusionadverse} that rely on synthetic data, highlighting the impact of real data for robustness. In the multi-frame setting, our method also surpasses the baseline \cite{manydepth}+\cite{gasperini_morbitzer2023md4all}, and even surpasses radar-based methods \cite{gasperini2021r4dyn}.
Tab. \ref{tab:robotcar} shows results on the Robotcar dataset \cite{robotcar} across daytime and nighttime domains. Since the test set contains only single-frame samples, we compare our method in the single-frame setting, achieving average improvements of $7.8\%, 6.5\%, 2.7\%$ on AbsRel, SqRel, and RMSE against recent methods \cite{gasperini_morbitzer2023md4all, diffusionadverse}. Notably, the baseline \cite{manydepth}+\cite{gasperini_morbitzer2023md4all} can not outperform previous single-frame methods. However, the model trained from our pipeline achieves noticeable improvements. \textbf{This indicates that the performance is not gained from the multi-frame backbone but from our training strategy.}

% Compared to the baseline \cite{manydepth}+\cite{gasperini_morbitzer2023md4all}, our method improves by $7.8\%, 20.7\%, 8.6\%$, maintaining the best $\delta_1$ performance, especially at night. 

We also evaluate different methods on the DrivingStereo dataset \cite{drivingstereo} (rain, fog) in a zero-shot setting, conducting in single-frame mode due to the lack of consecutive frames, as illustrated in Tab. \ref{tab:drivingstereo}. Ours surpasses the state-of-the-art by $3.0\%, 10.0\%, 2.8\%$ on AbsRel, SqRel, and RMSE, demonstrating stronger generalization ability over models trained on only synthetic data. Notably, our model significantly outperforms the fairly-designed baseline \cite{manydepth}+\cite{gasperini_morbitzer2023md4all} in both rain and fog conditions.
% These results confirm the robustness of our model across diverse conditions.

\subsection{Qualitative Experimental Results}
\label{sec:qualitative_exp}

Our qualitative results across various adverse conditions are shown in Fig.\ref{fig:qualitative_nuscenes_drivingstereo}. Due to space constraints, we display results from nuScenes and DrivingStereo; additional examples are provided in the supplementary materials. Since \cite{diffusionadverse} is not open source, we compare with the latest methods \cite{gasperini_morbitzer2023md4all, Saunders_2023_ICCV}.

Rows 1 and 2 show nuScenes examples under night and rain conditions. In row 1, the dark regions obscure details, yet the ground truth reveals a road sign on the left and trees on the right (highlighted in green). Robust-depth \cite{Saunders_2023_ICCV} and md4all \cite{gasperini_morbitzer2023md4all} struggle in these areas, missing the road sign and misestimating the depth of the trees. In contrast, our model provides reasonable depth predictions, closely aligning with the ground truth.
Row 2 illustrates a rainy scene with a blurry region partially obscuring a car. This blur causes \cite{gasperini_morbitzer2023md4all, Saunders_2023_ICCV} to overestimate depth, consistent with Fig.\ref{fig:analysis} (d). Our model, however, accurately identifies the nearby car, producing depth values close to the ground truth.
Row 3 shows a foggy scene from DrivingStereo, where previous methods misinterpret tire trails on the road as obstacles, while our model correctly ignores them, generating a smooth depth map. Row 4 presents a rainy DrivingStereo scene where reflections challenge even LiDAR depth capture, as shown in the green box. Both \cite{gasperini_morbitzer2023md4all, Saunders_2023_ICCV} are affected by reflections, but our model remains robust, providing depth consistent with geometric features. These examples demonstrate our model's adaptability to diverse real-world conditions.

\subsection{Analysis on Consistency Reweighting}
\begin{figure}[tb]
\centering
\includegraphics[width=0.98\linewidth, height=.2\linewidth]{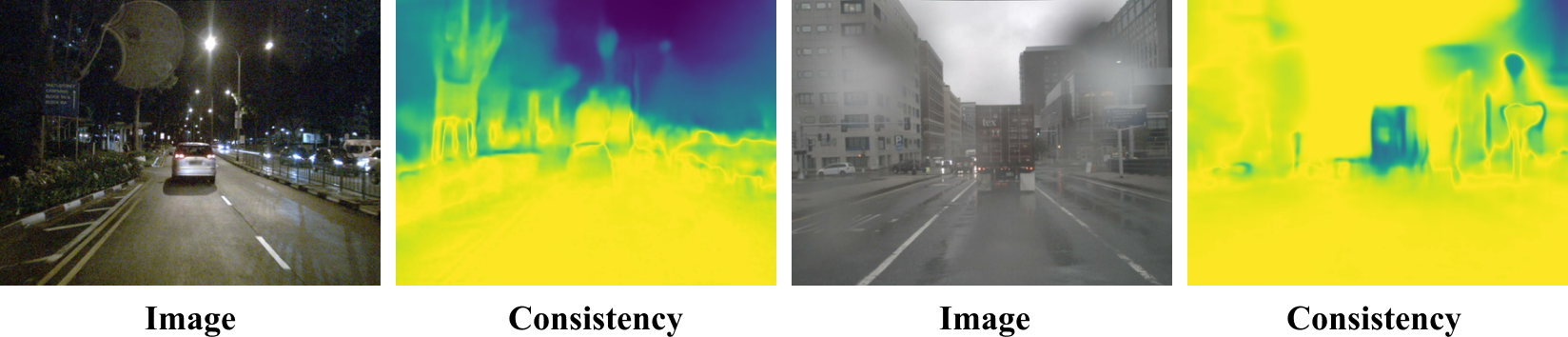}
\vspace{-0.1cm}
\caption{Visualization of consistency maps. }
\label{fig:consistency}
\vspace{-0.4cm}
\end{figure}

% The examples of consistency maps as discussed in Sec.\ref{method:real} is visualized in Fig.\ref{fig:consistency}, where brighter refers to large consistency. The first two columns correspond to the night condition, where it can be observed that the dark regions especially the distant parts usually have very weak consistency between $D_{day}$ and $D_{syn}$. On the contrary, the near regions own much stronger consistency. The last two rows refer to the rain condition, and it can be observed that the very inconsistent parts are those with severe blur which is easy to mislead the model. This gives us chances to emphasize the valid parts.
Examples of consistency maps discussed in Sec.\ref{method:real} are shown in Fig.\ref{fig:consistency}, where brighter areas indicate higher consistency. The first two columns correspond to night conditions, showing that dark, distant regions exhibit low consistency between $D_{day}$ and $D_{syn}$, while near regions have much stronger consistency. The last two columns represent rain conditions, where the blurry areas show low consistency. This allows us to select the more reliable regions. More visualization results can be seen in supplementary.

% \end{document}

\section{Ablations}
\label{sec:ablations}

% To verify the effectiveness of each stage with the designed modules, we conduct ablations on learning from cost volume($L_{cv}$), real adaptation without designed strategies(RA), consistency reweighting($L_{cd}$), sturcture prior constraint($L_{dis}^{KL}$). In the meanwhile, we compare learning from cost volume with direct learning from daytime feature($L_{feat}$). The baseline is built on a model only trained by depth loss in synthetic adaptation(SA). Due to the limited space, results for night and rain are shown in Tab.\ref{table:ablation}.

% %
% Due to the abundant information inside feature and the heterogeneous visual information in different conditions, simply using $L_{feat}$ drives the model with much worse performance compared with the baseline. On the contrary, since cost volume naturally filters other information and retain motion-structure feature, utilizing $L_{cv}$ boosts the model's performance in adverse scenarios. Moreover, directly apply RA without any designs can not be guaranteed to train a model work better in real data because not all supervisions are valid. However, by incorporating $L_{cd}$ for valid part emphasis and $L_{dis}^{KL}$ for structure prior constraint, the model's performance can be enhance gradually in the real-world adverse data. The ablation study verifies the effectiveness of different parts inside the pipeline the importance of synthetic-to-real learning.

To verify the effectiveness of each stage and the designed modules, we conduct ablations on learning from the cost volume ($L_{cv}$), real adaptation without designed strategies (RA), consistency reweighting (CR), and structure prior constraint ($L_{dis}^{KL}$). Additionally, we compare learning from the cost volume with direct learning from daytime features ($L_{feat}$). The baseline is trained solely with depth loss in the synthetic adaptation (SA). Due to space limitations, we provide results for night and rain conditions in Tab.\ref{table:ablation}.

Given the abundant information within features and the heterogeneous visual cues across conditions, simply using $L_{feat}$ leads to significantly worse performance than the baseline. In contrast, since the cost volume inherently filters extraneous information and retains motion-structure features, utilizing $L_{cv}$ enhances the model’s performance in adverse scenarios. Moreover, applying RA alone without additional strategies does not guarantee improved performance on real data, as not all supervisions are valid. However, by incorporating $L_{cd}$ to emphasize valid regions and $L_{dis}^{KL}$ for structural prior constraint, the model’s performance gradually improves in real-world adverse conditions. The ablation study confirms the effectiveness of the individual components within the pipeline and highlights the importance of synthetic-to-real adaptation.
\section{Conclusion}
\label{sec:conclusion}

% In this work, we elaborately design a synthetic-to-real learning pipeline for self-supervised robust depth estimation. It contains the synthetic adaptation and the real adaptation, where we construct the cost volume for daytime motion-structure feature transferring to adverse conditions. Moreover, the consistency reweighting strategy and structure prior constraint are proposed to boost the learning from synthetic to real domain. In experiments under different settings, our model can provide more accurate and robust depth estimation than the previous, and also generalizes better.

% In this work, we present a carefully designed synthetic-to-real learning pipeline for self-supervised robust depth estimation. Our approach consists of synthetic adaptation and real adaptation stages, in which we construct a cost volume to facilitate motion-structure feature transfer from daytime to adverse conditions. Additionally, we introduce a consistency reweighting strategy and a structural prior constraint to strengthen learning from the synthetic to the real domain. Experimental results across various settings demonstrate that our model achieves more accurate and robust depth estimation than previous methods, while also exhibiting superior generalization.
In this work, we propose a synthetic-to-real learning pipeline for self-supervised robust depth estimation, including synthetic adaptation and real adaptation stages where the cost volume is constructed to transfer motion-structure feature from daytime to adverse conditions. We introduce a consistency reweighting strategy and a structure prior constraint to boost learning from synthetic to real domains. Experiments across diverse settings and datasets show that our model provides more accurate and robust depth than previous methods, with superior generalization capabilities.

{
    \small
    \bibliographystyle{ieeenat_fullname}
    \bibliography{main}

\begin{thebibliography}{49}
\providecommand{\natexlab}[1]{#1}
\providecommand{\url}[1]{\texttt{#1}}
\expandafter\ifx\csname urlstyle\endcsname\relax
  \providecommand{\doi}[1]{doi: #1}\else
  \providecommand{\doi}{doi: \begingroup \urlstyle{rm}\Url}\fi

\bibitem[Avi-Aharon et~al.(2019)Avi-Aharon, Arbelle, and Raviv]{2019huenet}
Mor Avi-Aharon, Assaf Arbelle, and Tammy~Riklin Raviv.
\newblock Hue-net: Intensity-based image-to-image translation with differentiable histogram loss functions.
\newblock \emph{arXiv preprint arXiv:1912.06044}, 2019.

\bibitem[Avi-Aharon et~al.(2023)Avi-Aharon, Arbelle, and Raviv]{differentiable_hist}
Mor Avi-Aharon, Assaf Arbelle, and Tammy~Riklin Raviv.
\newblock Differentiable histogram loss functions for intensity-based image-to-image translation.
\newblock \emph{IEEE Transactions on Pattern Analysis and Machine Intelligence}, pages 1--12, 2023.

\bibitem[Bangunharcana et~al.(2023)Bangunharcana, Aly, and Kim]{bangunharcana2023dualrefine}
Antyanta Bangunharcana, Ahmed~Magd Aly, and KyungSoo Kim.
\newblock Dualrefine: Self-supervised depth and pose estimation through iterative epipolar sampling and refinement toward equilibrium.
\newblock In \emph{Proceedings of the IEEE/CVF Conference on Computer Vision and Pattern Recognition (CVPR)}. IEEE, 2023.

\bibitem[Bian et~al.(2021)Bian, Zhan, Wang, Li, Zhang, Shen, Cheng, and Reid]{bian2021ijcv_scdepth}
Jia-Wang Bian, Huangying Zhan, Naiyan Wang, Zhichao Li, Le Zhang, Chunhua Shen, Ming-Ming Cheng, and Ian Reid.
\newblock Unsupervised scale-consistent depth learning from video.
\newblock \emph{International Journal of Computer Vision (IJCV)}, 2021.

\bibitem[Caesar et~al.(2019)Caesar, Bankiti, Lang, Vora, Liong, Xu, Krishnan, Pan, Baldan, and Beijbom]{nuscenes2019}
Holger Caesar, Varun Bankiti, Alex~H. Lang, Sourabh Vora, Venice~Erin Liong, Qiang Xu, Anush Krishnan, Yu Pan, Giancarlo Baldan, and Oscar Beijbom.
\newblock nuscenes: A multimodal dataset for autonomous driving.
\newblock \emph{arXiv preprint arXiv:1903.11027}, 2019.

\bibitem[Feng et~al.(2022)Feng, Yang, Jing, Wang, Tian, and Li]{disentangling}
Ziyue Feng, Liang Yang, Longlong Jing, Haiyan Wang, YingLi Tian, and Bing Li.
\newblock Disentangling object motion and occlusion for unsupervised multi-frame monocular depth.
\newblock \emph{arXiv preprint arXiv:2203.15174}, 2022.

\bibitem[Gasperini et~al.(2021)Gasperini, Koch, Dallabetta, Navab, Busam, and Tombari]{gasperini2021r4dyn}
Stefano Gasperini, Patrick Koch, Vinzenz Dallabetta, Nassir Navab, Benjamin Busam, and Federico Tombari.
\newblock R4dyn: Exploring radar for self-supervised monocular depth estimation of dynamic scenes.
\newblock In \emph{2021 International Conference on 3D Vision (3DV)}, pages 751--760. IEEE, 2021.

\bibitem[Gasperini et~al.(2023)Gasperini, Morbitzer, Jung, Navab, and Tombari]{gasperini_morbitzer2023md4all}
Stefano Gasperini, Nils Morbitzer, HyunJun Jung, Nassir Navab, and Federico Tombari.
\newblock Robust monocular depth estimation under challenging conditions.
\newblock In \emph{Proceedings of the IEEE/CVF International Conference on Computer Vision}, pages 8177--8186, 2023.

\bibitem[Godard et~al.(2017)Godard, Aodha, and Brostow]{2017left-right-consistency}
Clément Godard, Oisin~Mac Aodha, and Gabriel~J. Brostow.
\newblock Unsupervised monocular depth estimation with left-right consistency.
\newblock In \emph{2017 IEEE Conference on Computer Vision and Pattern Recognition (CVPR)}, pages 6602--6611, 2017.

\bibitem[Godard et~al.(2019)Godard, Mac~Aodha, Firman, and Brostow]{godard2019digging}
Cl{\'e}ment Godard, Oisin Mac~Aodha, Michael Firman, and Gabriel~J Brostow.
\newblock Digging into self-supervised monocular depth estimation.
\newblock In \emph{Proceedings of the IEEE/CVF international conference on computer vision}, pages 3828--3838, 2019.

\bibitem[Guizilini et~al.(2020)Guizilini, Ambrus, Pillai, Raventos, and Gaidon]{packnet-sfm}
Vitor Guizilini, Rares Ambrus, Sudeep Pillai, Allan Raventos, and Adrien Gaidon.
\newblock 3d packing for self-supervised monocular depth estimation.
\newblock In \emph{IEEE Conference on Computer Vision and Pattern Recognition (CVPR)}, 2020.

\bibitem[Guizilini et~al.(2022)Guizilini, Ambruș, Chen, Zakharov, and Gaidon]{multi_frame_transformer}
Vitor Guizilini, Rareș Ambruș, Dian Chen, Sergey Zakharov, and Adrien Gaidon.
\newblock Multi-frame self-supervised depth with transformers.
\newblock In \emph{Proceedings of the IEEE/CVF Conference on Computer Vision and Pattern Recognition (CVPR)}, pages 160--170, 2022.

\bibitem[He et~al.(2016)He, Zhang, Ren, and Sun]{resnet}
Kaiming He, Xiangyu Zhang, Shaoqing Ren, and Jian Sun.
\newblock Deep residual learning for image recognition.
\newblock In \emph{2016 IEEE Conference on Computer Vision and Pattern Recognition (CVPR)}, pages 770--778, 2016.

\bibitem[Li et~al.(2021)Li, Huang, and Zhang]{Liming_2021_ICCV}
Ming Li, Xinming Huang, and Ziming Zhang.
\newblock Self-supervised geometric features discovery via interpretable attention for vehicle re-identification and beyond.
\newblock In \emph{Proceedings of the IEEE/CVF International Conference on Computer Vision (ICCV)}, pages 194--204, 2021.

\bibitem[Li et~al.(2023{\natexlab{a}})Li, Liu, Zheng, Huang, and Zhang]{LiMingTMM2021}
Ming Li, Jun Liu, Ce Zheng, Xinming Huang, and Ziming Zhang.
\newblock Exploiting multi-view part-wise correlation via an efficient transformer for vehicle re-identification.
\newblock \emph{IEEE Transactions on Multimedia}, 25:\penalty0 919--929, 2023{\natexlab{a}}.

\bibitem[Li et~al.(2023{\natexlab{b}})Li, Xu, Fan, Zhou, Liu, Liu, Li, Keppo, Shou, and Yan]{Li_2023_ICCV}
Ming Li, Xiangyu Xu, Hehe Fan, Pan Zhou, Jun Liu, Jia-Wei Liu, Jiahe Li, Jussi Keppo, Mike~Zheng Shou, and Shuicheng Yan.
\newblock Stprivacy: Spatio-temporal privacy-preserving action recognition.
\newblock In \emph{Proceedings of the IEEE/CVF International Conference on Computer Vision (ICCV)}, pages 5106--5115, 2023{\natexlab{b}}.

\bibitem[Li et~al.(2023{\natexlab{c}})Li, Zhou, Liu, Keppo, Lin, Yan, and Xu]{li2023instant3d}
Ming Li, Pan Zhou, Jia-Wei Liu, Jussi Keppo, Min Lin, Shuicheng Yan, and Xiangyu Xu.
\newblock Instant3d: Instant text-to-3d generation.
\newblock \emph{arxiv: 2311.08403}, 2023{\natexlab{c}}.

\bibitem[Liu et~al.(2021)Liu, Song, Wang, Liu, and Zhang]{liu2021ADIDS}
Lina Liu, Xibin Song, Mengmeng Wang, Yong Liu, and Liangjun Zhang.
\newblock Self-supervised monocular depth estimation for all day images using domain separation.
\newblock In \emph{Proceedings of the IEEE/CVF International Conference on Computer Vision}, pages 12737--12746, 2021.

\bibitem[Maddern et~al.(2016)Maddern, Pascoe, Linegar, and Newman]{robotcar}
W. Maddern, G. Pascoe, C. Linegar, and P. Newman.
\newblock 1 year, 1000 km: The oxford robotcar dataset.
\newblock \emph{International Journal of Robotics Research}, page 0278364916679498, 2016.

\bibitem[Peng et~al.(2021)Peng, Wang, Lai, Tang, and Cai]{2021epcdepth}
Rui Peng, Ronggang Wang, Yawen Lai, Luyang Tang, and Yangang Cai.
\newblock Excavating the potential capacity of self-supervised monocular depth estimation.
\newblock In \emph{Proceedings of the IEEE International Conference on Computer Vision (ICCV)}, 2021.

\bibitem[Rombach et~al.(2022)Rombach, Blattmann, Lorenz, Esser, and Ommer]{LDM}
Robin Rombach, Andreas Blattmann, Dominik Lorenz, Patrick Esser, and Bj\"orn Ommer.
\newblock High-resolution image synthesis with latent diffusion models.
\newblock In \emph{Proceedings of the IEEE/CVF Conference on Computer Vision and Pattern Recognition (CVPR)}, pages 10684--10695, 2022.

\bibitem[Saunders et~al.(2023)Saunders, Vogiatzis, and Manso]{Saunders_2023_ICCV}
Kieran Saunders, George Vogiatzis, and Luis~J. Manso.
\newblock Self-supervised monocular depth estimation: Let's talk about the weather.
\newblock In \emph{Proceedings of the IEEE/CVF International Conference on Computer Vision (ICCV)}, pages 8907--8917, 2023.

\bibitem[Spencer et~al.(2020)Spencer, Bowden, and Hadfield]{spencer2020defeat}
Jaime Spencer, Richard Bowden, and Simon Hadfield.
\newblock Defeat-net: General monocular depth via simultaneous unsupervised representation learning.
\newblock In \emph{Proceedings of the IEEE/CVF Conference on Computer Vision and Pattern Recognition}, pages 14402--14413, 2020.

\bibitem[Sun et~al.(2023)Sun, Bian, Zhan, Yin, Reid, and Shen]{sc_depthv3}
Libo Sun, Jia-Wang Bian, Huangying Zhan, Wei Yin, Ian Reid, and Chunhua Shen.
\newblock Sc-depthv3: Robust self-supervised monocular depth estimation for dynamic scenes.
\newblock \emph{IEEE Transactions on Pattern Analysis and Machine Intelligence (TPAMI)}, 2023.

\bibitem[Tosi et~al.(2024)Tosi, Zama~Ramirez, and Poggi]{diffusionadverse}
Fabio Tosi, Pierluigi Zama~Ramirez, and Matteo Poggi.
\newblock Diffusion models for monocular depth estimation: Overcoming challenging conditions.
\newblock In \emph{European Conference on Computer Vision (ECCV)}, 2024.

\bibitem[Vankadari et~al.(2020)Vankadari, Garg, Majumder, Kumar, and Behera]{vankadari2020ADFA}
Madhu Vankadari, Sourav Garg, Anima Majumder, Swagat Kumar, and Ardhendu Behera.
\newblock Unsupervised monocular depth estimation for night-time images using adversarial domain feature adaptation.
\newblock In \emph{Computer Vision--ECCV 2020: 16th European Conference, Glasgow, UK, August 23--28, 2020, Proceedings, Part XXVIII 16}, pages 443--459. Springer, 2020.

\bibitem[Vankadari et~al.(2023)Vankadari, Golodetz, Garg, Shin, Markham, and Trigoni]{vankadari2023sun}
Madhu Vankadari, Stuart Golodetz, Sourav Garg, Sangyun Shin, Andrew Markham, and Niki Trigoni.
\newblock When the sun goes down: Repairing photometric losses for all-day depth estimation.
\newblock In \emph{Conference on Robot Learning}, pages 1992--2003. PMLR, 2023.

\bibitem[Wang et~al.(2023{\natexlab{a}})Wang, Lin, Nie, Huang, Zhao, Pan, and Ai]{wang2023weatherdepth}
Jiyuan Wang, Chunyu Lin, Lang Nie, Shujun Huang, Yao Zhao, Xing Pan, and Rui Ai.
\newblock Weatherdepth: Curriculum contrastive learning for self-supervised depth estimation under adverse weather conditions, 2023{\natexlab{a}}.

\bibitem[Wang et~al.(2024)Wang, Nie, Liao, Shao, and Zhao]{diffusion_contrast}
Jiyuan Wang, Lang Nie, Kang Liao, Shuwei Shao, and Yao Zhao.
\newblock Digging into contrastive learning for robust depth estimation with diffusion models.
\newblock In \emph{ACM Int. Conf. Multimedia (ACMMM)}, pages 4129--4137, 2024.

\bibitem[Wang et~al.(2021)Wang, Zhang, Yan, Li, Xu, Li, and Yang]{wang2021RNW}
Kun Wang, Zhenyu Zhang, Zhiqiang Yan, Xiang Li, Baobei Xu, Jun Li, and Jian Yang.
\newblock Regularizing nighttime weirdness: Efficient self-supervised monocular depth estimation in the dark.
\newblock In \emph{Proceedings of the IEEE/CVF International Conference on Computer Vision}, pages 16055--16064, 2021.

\bibitem[Wang et~al.(2023{\natexlab{b}})Wang, Yu, and Gao]{2023planedepth}
Ruoyu Wang, Zehao Yu, and Shenghua Gao.
\newblock Planedepth: Self-supervised depth estimation via orthogonal planes.
\newblock In \emph{2023 IEEE/CVF Conference on Computer Vision and Pattern Recognition (CVPR)}, pages 21425--21434, 2023{\natexlab{b}}.

\bibitem[Wang et~al.(2023{\natexlab{c}})Wang, Liang, Xu, Jiao, and Yu]{wang2023sqldepth}
Youhong Wang, Yunji Liang, Hao Xu, Shaohui Jiao, and Hongkai Yu.
\newblock Sqldepth: Generalizable self-supervised fine-structured monocular depth estimation, 2023{\natexlab{c}}.

\bibitem[Watson et~al.(2021)Watson, Mac~Aodha, Prisacariu, Brostow, and Firman]{manydepth}
Jamie Watson, Oisin Mac~Aodha, Victor Prisacariu, Gabriel Brostow, and Michael Firman.
\newblock The temporal opportunist: Self-supervised multi-frame monocular depth.
\newblock In \emph{Proceedings of the IEEE/CVF Conference on Computer Vision and Pattern Recognition (CVPR)}, pages 1164--1174, 2021.

\bibitem[Woo et~al.(2024)Woo, Lee, Kim, Lee, and Lee]{prodepth}
Sungmin Woo, Wonjoon Lee, Woo~Jin Kim, Dogyoon Lee, and Sangyoun Lee.
\newblock Prodepth: Boosting self-supervised multi-frame monocular depth with probabilistic fusion.
\newblock \emph{arXiv preprint arXiv:2407.09303}, 2024.

\bibitem[Wu et~al.(2024)Wu, Cao, Li, Chen, Wang, Meng, and Huang]{Wu2024TowardsBT}
Yihang Wu, Xiao Cao, Kaixin Li, Zitan Chen, Haonan Wang, Lei Meng, and Zhiyong Huang.
\newblock Towards better text-to-image generation alignment via attention modulation.
\newblock \emph{ArXiv}, abs/2404.13899, 2024.

\bibitem[Xu et~al.(2023)Xu, Zhang, Cai, Rezatofighi, Yu, Tao, and Geiger]{xu2023unifying}
Haofei Xu, Jing Zhang, Jianfei Cai, Hamid Rezatofighi, Fisher Yu, Dacheng Tao, and Andreas Geiger.
\newblock Unifying flow, stereo and depth estimation.
\newblock \emph{IEEE Transactions on Pattern Analysis and Machine Intelligence}, 2023.

\bibitem[Yang et~al.(2019)Yang, Song, Huang, Deng, Shi, and Zhou]{drivingstereo}
Guorun Yang, Xiao Song, Chaoqin Huang, Zhidong Deng, Jianping Shi, and Bolei Zhou.
\newblock Drivingstereo: A large-scale dataset for stereo matching in autonomous driving scenarios.
\newblock In \emph{IEEE Conference on Computer Vision and Pattern Recognition (CVPR)}, 2019.

\bibitem[Zhang et~al.(2023{\natexlab{a}})Zhang, Rao, and Agrawala]{controlnet}
Lvmin Zhang, Anyi Rao, and Maneesh Agrawala.
\newblock Adding conditional control to text-to-image diffusion models.
\newblock In \emph{IEEE International Conference on Computer Vision (ICCV)}, 2023{\natexlab{a}}.

\bibitem[Zhang et~al.(2023{\natexlab{b}})Zhang, Nex, Vosselman, and Kerle]{litemono}
Ning Zhang, Francesco Nex, George Vosselman, and Norman Kerle.
\newblock Lite-mono: A lightweight cnn and transformer architecture for self-supervised monocular depth estimation.
\newblock In \emph{Proceedings of the IEEE/CVF Conference on Computer Vision and Pattern Recognition (CVPR)}, pages 18537--18546, 2023{\natexlab{b}}.

\bibitem[Zhang et~al.(2024)Zhang, Xie, Yuan, Mi, and Tan]{xin2024heap}
Xin Zhang, Jinheng Xie, Yuan Yuan, Michael~Bi Mi, and Robby~T Tan.
\newblock Heap: unsupervised object discovery and localization with contrastive grouping.
\newblock In \emph{Proceedings of the AAAI Conference on Artificial Intelligence}, pages 7323--7331, 2024.

\bibitem[Zhao et~al.(2022{\natexlab{a}})Zhao, Tang, and Sun]{zhao2022ITDFA}
Chaoqiang Zhao, Yang Tang, and Qiyu Sun.
\newblock Unsupervised monocular depth estimation in highly complex environments.
\newblock \emph{IEEE Transactions on Emerging Topics in Computational Intelligence}, 6\penalty0 (5):\penalty0 1237--1246, 2022{\natexlab{a}}.

\bibitem[Zhao et~al.(2022{\natexlab{b}})Zhao, Zhang, Poggi, Tosi, Guo, Zhu, Huang, Tang, and Mattoccia]{zhao2022monovit}
Chaoqiang Zhao, Youmin Zhang, Matteo Poggi, Fabio Tosi, Xianda Guo, Zheng Zhu, Guan Huang, Yang Tang, and Stefano Mattoccia.
\newblock Monovit: Self-supervised monocular depth estimation with a vision transformer.
\newblock In \emph{2022 International Conference on 3D Vision (3DV)}, pages 668--678. IEEE, 2022{\natexlab{b}}.

\bibitem[Zheng et~al.(2020{\natexlab{a}})Zheng, Lyu, Li, and Zhang]{LodoNet}
Ce Zheng, Yecheng Lyu, Ming Li, and Ziming Zhang.
\newblock Lodonet: A deep neural network with 2d keypoint matching for 3d lidar odometry estimation.
\newblock In \emph{Proceedings of the 28th ACM International Conference on Multimedia}, page 2391–2399, New York, NY, USA, 2020{\natexlab{a}}. Association for Computing Machinery.

\bibitem[Zheng et~al.(2023)Zheng, Zhong, Li, Gao, Zheng, Jin, Wang, Zhao, Zhou, Zhang, and Zhao]{23ICRA_Steps}
Yupeng Zheng, Chengliang Zhong, Pengfei Li, Huan-ang Gao, Yuhang Zheng, Bu Jin, Ling Wang, Hao Zhao, Guyue Zhou, Qichao Zhang, and Dongbin Zhao.
\newblock Steps: Joint self-supervised nighttime image enhancement and depth estimation.
\newblock In \emph{2023 IEEE International Conference on Robotics and Automation (ICRA)}, pages 4916--4923, 2023.

\bibitem[Zheng et~al.(2020{\natexlab{b}})Zheng, Wu, Han, and Shi]{zheng_2020_forkgan}
Ziqiang Zheng, Yang Wu, Xinran Han, and Jianbo Shi.
\newblock Forkgan: Seeing into the rainy night.
\newblock In \emph{The IEEE European Conference on Computer Vision (ECCV)}, 2020{\natexlab{b}}.

\bibitem[Zhou et~al.(2023)Zhou, Chang, Yan, and Yan]{UCDA}
Hanyu Zhou, Yi Chang, Wending Yan, and Luxin Yan.
\newblock Unsupervised cumulative domain adaptation for foggy scene optical flow.
\newblock In \emph{Proceedings of the IEEE/CVF Conference on Computer Vision and Pattern Recognition}, pages 9569--9578, 2023.

\bibitem[Zhou et~al.(2024)Zhou, Chang, Liu, WENDING, Duan, Shi, and Yan]{zhou2024exploring}
Hanyu Zhou, Yi Chang, Haoyue Liu, YAN WENDING, Yuxing Duan, Zhiwei Shi, and Luxin Yan.
\newblock Exploring the common appearance-boundary adaptation for nighttime optical flow.
\newblock In \emph{The Twelfth International Conference on Learning Representations}, 2024.

\bibitem[Zhou et~al.(2017)Zhou, Brown, Snavely, and Lowe]{2017cvpr_egomotion}
Tinghui Zhou, Matthew Brown, Noah Snavely, and David~G. Lowe.
\newblock Unsupervised learning of depth and ego-motion from video.
\newblock In \emph{2017 IEEE Conference on Computer Vision and Pattern Recognition (CVPR)}, pages 6612--6619, 2017.

\bibitem[Zhu et~al.(2017)Zhu, Park, Isola, and Efros]{CycleGAN2017}
Jun-Yan Zhu, Taesung Park, Phillip Isola, and Alexei~A Efros.
\newblock Unpaired image-to-image translation using cycle-consistent adversarial networkss.
\newblock In \emph{Computer Vision (ICCV), 2017 IEEE International Conference on}, 2017.

\end{thebibliography}
}

% WARNING: do not forget to delete the supplementary pages from your submission 
% \input{sec/X_suppl}

\end{document}

\maketitle

In the supplementary of the main text in the paper, we illustrate the following contents for better understanding the proposed method: (1) More implementation details for our experiments in Sec.\ref{supp:implementation_details}. (2) The complete differentiable histogram formula and comparison with GAN in Sec.\ref{supp:differentiable_histogram}. (3) More qualitative results on nuScenes, Robotcar and DrivingStereo \cite{nuscenes2019,robotcar,drivingstereo} in Sec.\ref{supp:more_qualitative_results}. (4) More visualization examples of the consistency maps in Sec.\ref{supp:consistency}.

\section{Implementation Details}
\label{supp:implementation_details}
The experiments on all models of different stages are conducted on the same ResNet architecture \cite{resnet}, so do the baselines. The training for the $\Phi_{day}$ and the $\Phi_{syn}$ are conducted for 20 epochs, while $\Phi_{real}$ is initialized by $\Phi_{syn}$ and trained for 10 epochs. The learning rate is initialized to $1\times 10^{-4}$ with a decreasing factor of 0.5 every 5 epochs. For the adaptive depth bins in Manydepth \cite{manydepth}, we fix them to $3.5-80m$ since \cite{gasperini_morbitzer2023md4all, diffusionadverse, packnet-sfm} all use a weak velocity loss to align the scales. Thanks to \cite{gasperini_morbitzer2023md4all, diffusionadverse}, we can utilize the synthetic data from their pretrained augmentation models. For the $Dist_{day}$, we obtain a fixed distribution through the whole daytime training dataset, and utilize it for the real adaptation constraint. The training of all the pipeline is conducted on a single NVIDIA A5000 GPU. The synthetic adaptation takes about 14 hours of training, while the real adaptation takes about 7 hours for training.

For the training objective of synthetic adaptation $L_{syn}$, we simply set $\alpha_1=1.0, \alpha_2=1.0, \alpha_3=1.0$; while for the total objective in real adaptation, we set $\alpha_1=1.0, \alpha_2=0.01, \alpha_3=1.0, \alpha_4=1.0$. For the consistency-reweighting strategy, the scale factor $\beta$ and the weight bias $\epsilon$ are chosen to be $1.0, 1.0$ respectively.

\noindent \textbf{Pose Supervision.} Since pose is crucial to construct a cost volume, we also use the daytime pose model to supervise the pose model for synthetic adverse conditions with L2 loss. Specifically, we utilize the rotation matrix represented by the angles, where we have $\theta_{day}$ and $\theta_{syn}$; and the translation vectors: $t_{day}, t_{syn}$. The supervision for pose from daytime to adverse conditions is written as
\begin{align}
    L_\theta =& \Vert \theta_{day}-\theta_{syn} \Vert_2, \\
    L_t =& \Vert t_{day}-t_{syn} \Vert_2, \\
    L_T =& L_\theta + L_t.
\end{align}

The pose objective in real adaptation is similar to the above equations.

\noindent \textbf{The Proportion of Data in Different Conditions.} In the synthetic adaptation, we set the proportion of daytime, nighttime and rain to be $50\%, 25\%, 25\%$. This is because training the model in multi-frame mode and single-frame augmentation are quite important, and this leads us to set more daytime data in training. In the real adaptation, we utilize the data of multiple conditions inside dataset without any specific design for the proportion.

\begin{table*}[htb] 
\renewcommand{\arraystretch}{1.2}

\centering
% \Huge
\footnotesize
% \small
\setlength{\tabcolsep}{4pt}  % 调整列间距
\caption{Quantitative comparison between using GAN and using our proposed structure prior. Using structure prior can further decrease the gaps between adverse conditions and daytime, while it is difficult to narrow this fine gap using GAN.}
\begin{tabular}{l|c|cccc|cccc|cccc} %需要10列
\toprule %添加表格头部粗线
\cline{1-14}\
\multirow{2}{*}{Method} & \multicolumn{1}{c|}{Test} & \multicolumn{4}{c|}{nuScenes-day} & \multicolumn{4}{c|}{nuScenes-night} & \multicolumn{4}{c}{nuScenes-rain}\\
     &frames& AbsRel$\downarrow$ & SqRel$\downarrow$ &  RMSE$\downarrow$ & $ \delta_1\uparrow $ & AbsRel$\downarrow$ & SqRel$\downarrow$ &  RMSE$\downarrow$ & $ \delta_1\uparrow $ & AbsRel$\downarrow$ & SqRel$\downarrow$ &  RMSE$\downarrow$ & $ \delta_1\uparrow $    \\  
% \midrule
\cline{1-14}

GAN & 2(-1,0) &0.1234 &1.446 &6.069 &86.42 &0.1761 &1.841 &7.823 &72.37&0.1332 &1.614 &6.766 &82.20 \\
\textbf{Structure Prior} & 2(-1,0) &\textbf{0.1170} &\textbf{1.371} &\textbf{6.023} &\textbf{86.66} & \textbf{0.1713} & \textbf{1.779} &\textbf{7.689} &\textbf{73.16} &\textbf{0.1266} &\textbf{1.532} &\textbf{6.654} &\textbf{82.86} \\
\cline{1-14}
\bottomrule 
\end{tabular}
\label{tab:compare_GAN_prior}
% \vspace{-0.2cm}
\end{table*}

\section{Differentiable Histogram}
\label{supp:differentiable_histogram}
The target of differentiable histogram is to using a differentiable function to approximate the step function of the histogram statistical process. We follow the design in \cite{2019huenet, differentiable_hist}, using the Kernel Density Estimation (KDE) with a kernel $K$ to estimate the intensity density $f_D$ of a depth map $D_{daytime}$ or $D_{adverse}$:
\begin{align}
    f_D(d) = \frac{1}{hwa} \sum_{x \in \Omega} K\left( \frac{D(x) - d}{a} \right),
\end{align}
where $d$ represents the depth value, $\Omega$ is the pixels on a depth map, $a$ is the bandwidth, and $h,w$ refer to the size of map. The kernel $K$ is chosen as the derivative of the sigmoid function:
\begin{align}
    K(x) =& \frac{d}{dx} \sigma(x) = \sigma(x) \sigma(-x), \label{equ:kernel} \\
    \sigma&(x) = \frac{1}{1 + e^{-x}}.
\end{align}

From Equ.\ref{equ:kernel}, the kernel is obvious to have non-negative, symmetric characteristics, which is consistent with the requirements of KDE. Then, we compute the depth map of range $d_{min}~d_{max}$ to a differentiable histogram by dividing the depth into N sub-intervals. The $n^{th}$ interval owns the length of $L=\frac{d_{max}-d_{min}}{N}$ and the center $b_n=d_{min}+L(n+\frac{1}{2})$. Our target is to obtain the probability of one pixel on the depth map belonging to the $n^{th}$ sub-interval, $P(n)$:
\begin{align}
    P(n) = \int_{nL+d_{min}}^{(n+1)L+d_{min}} f_D(x)dx.
\end{align}

After solving the integration, $P(n)$ can be rewritten as
\begin{align}
     P(n) =\frac{1}{hw}\sum_{x=1}^{hw}[&\sigma(\frac{D(x)-b_n+\frac{L}{2}}{a}) \notag\\
     &-\sigma(\frac{D(x)-b_n-\frac{L}{2}}{a})],
     \label{equ:differential_histogram}
\end{align}

By calculating across every depth sub-interval, the histogram can be formulated as: 
\begin{align}
     hist =  \{b_n, P(n)\}_{n=1}^{N},\label{equ:distribution_day}
\end{align}

Computing through all the real-world data, we can have $hist_{day}, hist_{night}, hist_{rain}$ for separate conditions. $hist_{day}$ corresponds to $Dist_{day}$ in our main paper, which is utilized as a reference distribution guidance.

In our experiments, $d_{min}=3.5, d_{max}=80.0, N=100, a=\frac{L}{20}$ to approximate the step function. Noted that the reference distribution is calculated through the whole training set of daytime data with a frozen daytime model in our experiments.

\noindent \textbf{Distribution Prior v.s. GAN.} What's more, comparing with the utility of GAN \cite{wang2021RNW}, we notice two most important advantages of this explicit differentiable distribution: (1) Leveraging GAN requires to randomly sample a batch of daytime data in every iteration, which can not guarantee the quality of the sampled data and possible to lead the models to be optimized with randomness. Our design of the explicit distribution is calculated through the dataset of daytime data, which is much more representative and stable for optimization. (2) Using GAN is proper or meaningful when the gaps are quite huge (like optimize a daytime model to directly estimate in nighttime). However, when the gaps have decreased significantly (after training a model in synthetic adverse conditions with pseudo-labels), it is hard for GAN to close the gaps further, due to the rough supervision and optimization mechanism of GAN (as in our experiments, the discriminator is hard to decrease its loss, because the domain gap between current prediction and daytime prediction is much smaller than not doing synthetic adaptation). On the contrary, we find that a fine distribution of statistical depth can further help to decrease the gaps, and the comparison between utility of structure prior and GAN is shown in Tab.\ref{tab:compare_GAN_prior}.

\section{Qualitative Results on Different Datasets}
\label{supp:more_qualitative_results}
As supplementary to the main paper, we provide more qualitative results on nuScenes \cite{nuscenes2019} (daytime, nighttime, rain), Robotcar \cite{robotcar} (daytime, nighttime) and DrivingStereo \cite{drivingstereo} (rain, fog), as shown from Fig.\ref{fig:qualitative_nuscenes_daytime}-\ref{fig:qualitative_drivingstereo_fog}.

\subsection{NuScenes}

Our model, as shown in Fig.\ref{fig:qualitative_nuscenes_daytime}, provides comparative daytime predictions as the state-of-the-art, and from Tab.1 in the main paper, ours actually estimates more accurately, which is due to our strategy to transfer motion-structure knowledge for better robust representation learning.

Fig.\ref{fig:qualitative_nuscenes_nighttime} displays more examples in the nighttime condition, which is quite severe due to its extreme darkness and glare. In row 2 to row 3, previous methods all fail to precisely capture depth of the left and parts (where we can see tree on both sides from the ground truth), while our model gives accurate predictions on both left and right trees. In row 4 to row 7, it can be observed from the ground truth that there is a tree on the left of the road and very close to the camera, while \cite{gasperini_morbitzer2023md4all, Saunders_2023_ICCV} detect the objects as far away. However, ours gives much more accurate predictions. In row 8 and 9, previous method suffer from the glare of blinding headlights, giving wrong predictions especially on the right side of the image. On the contrary, ours provides natural and correct depth to the glare parts. In row 10, other methods predict the tree which is far on the left of the view as much closer, and that is highly inconsistent with the ground truth. Ours captures this character and gives reasonable results.

Fig.\ref{fig:qualitative_nuscenes_rain} refers to more visualization examples of the rain condition in nuScenes dataset. From row 1 to row 7, all previous methods are significantly affected by the obvious reflections and blur which are common in rain weather. For instance, in row 2 there is a car with severe blur on the left of the image, where both \cite{gasperini_morbitzer2023md4all} and \cite{Saunders_2023_ICCV} both fail and give incorrect results to this part. Ours as seen in row 2, column 5, displays much more robust and precise depth to the challenging region. In row 4, the previous approaches are misguided by the reflections of the white truck to perceive the reflections as obstacles on the road, while our model gives much more smooth depth map of the road. Considering row 9 and 10, we surprisingly found that our model can even less influenced by the moving objects in a scene, compared with \cite{gasperini_morbitzer2023md4all}. This is due to the consistency-reweighting strategy which can assign smaller supervision weights to the corresponding regions.

\subsection{Robotcar}

In Fig.\ref{fig:qualitative_robotcar}, we display more qualitative results in daytime and nighttime conditions. Our model performs similarly in daytime, while significantly provides more robust and accurate estimations in nighttime. In row 3, there is a double-decker bus on the right, where the second deck can not be perceived by \cite{gasperini_morbitzer2023md4all} due to its darkness, and our method is capable of recognizing this part. In row 4, a traffic light on the right side is affected by both darkness and glare, which make the previous approach unable to predict its depth. However, ours also successfully provide correct depth to the light.

\subsection{DrivingStereo}
From Fig.\ref{fig:qualitative_drivingstereo_rain} to Fig.\ref{fig:qualitative_drivingstereo_fog}, more examples of DrivingStereo dataset \cite{drivingstereo} in rain and fog conditions are listed. In Fig.\ref{fig:qualitative_drivingstereo_rain}, md4all \cite{gasperini_morbitzer2023md4all} and robust-depth \cite{Saunders_2023_ICCV} are obviously affected by the reflections, such as perceiving the reflections as severe obstacles on the roads (row 3, row 6, row 8-10). In some scenarios the model can not distinguish the objects with its reflections (row 4), leading to wrong predictions. On the contrary, ours is consistently robust to the degradation.

In Fig.\ref{fig:qualitative_drivingstereo_fog}, there are more examples in the fog conditions. In short, previous methods \cite{gasperini_morbitzer2023md4all, Saunders_2023_ICCV} easily suffer from the noises like traces or reflections on the road. For example, in row 6 there are obvious reflections of the roadside sign on the right, which affects other methods' estimation to the road. Ours, however, is not affected by such reflections. Another example can be seen in row 9, where previous methods are influenced by the windows of the bus (left side of row 9), resulting in erroneous predictions of the bus. Compared with those results, our model gives smoother depth to the whole bus.

\section{Consistency Map Visualization}
\label{supp:consistency}
Our consistency reweighting strategy aims at dynamically assigning importance to different regions of the input based on the agreement between multiple models' predictions, because there is no perfect ground truth.
To verify the effectiveness of the proposed consistency-reweighitng strategy, we list more visualization examples in challenging conditions in Fig.\ref{fig:viz_consistency}. Considering the consistency maps, the brighter represents the more consistent, and the darker means the more inconsistent. Notably, for night condition, it can be observed the inconsistency is resulted mainly from the extreme dark regions or the parts with severe glare. For the regions with good visibility (such as the near road), it is always consistent. In rain condition, the inconsistency is usually lead by the blur or the reflections. By introducing the consistency maps, we target to provide strong supervision to the reliable regions, such as clear roads with good visibility, and conduct weak supervision to the challenging regions in real-world data. This is important because our earlier trained models can not be guaranteed to provide perfect pseudo-depth labels when facing real world adverse conditions.

\begin{figure*}[!t]
    \centering
    % 第一部分: nuScenes
\begin{tabular}{c @{} c @{} c  @{} c @{} c @{} c }
        &
        \includegraphics[width=0.19\textwidth]{figures/figures_supp/nuscenes/daytime/n008-2018-08-01-15-16-36-0400__CAM_FRONT__1533151373612404.jpg} \hspace{1pt} &
        \includegraphics[width=0.19\textwidth]{figures/figures_supp/nuscenes/daytime/n008-2018-08-01-15-16-36-0400__CAM_FRONT__1533151373612404_depth_gt.jpg} \hspace{1pt} &
        \includegraphics[width=0.19\textwidth]{figures/figures_supp/nuscenes/daytime/n008-2018-08-01-15-16-36-0400__CAM_FRONT__1533151373612404_inv_depth_robust_depth.jpg} \hspace{1pt} &
        \includegraphics[width=0.19\textwidth]{figures/figures_supp/nuscenes/daytime/n008-2018-08-01-15-16-36-0400__CAM_FRONT__1533151373612404_inv_depth_md4all.jpg} \hspace{1pt} &
        \includegraphics[width=0.19\textwidth]{figures/figures_supp/nuscenes/daytime/n008-2018-08-01-15-16-36-0400__CAM_FRONT__1533151373612404_inv_depth_ours.jpg} \\
        
        %  row 2
        &
        \includegraphics[width=0.19\textwidth]{figures/figures_supp/nuscenes/daytime/n008-2018-08-01-15-16-36-0400__CAM_FRONT__1533151477612404.jpg} \hspace{1pt} &
        \includegraphics[width=0.19\textwidth]{figures/figures_supp/nuscenes/daytime/n008-2018-08-01-15-16-36-0400__CAM_FRONT__1533151477612404_depth_gt.jpg} \hspace{1pt} &
        \includegraphics[width=0.19\textwidth]{figures/figures_supp/nuscenes/daytime/n008-2018-08-01-15-16-36-0400__CAM_FRONT__1533151477612404_inv_depth_robust_depth.jpg} \hspace{1pt} &
        \includegraphics[width=0.19\textwidth]{figures/figures_supp/nuscenes/daytime/n008-2018-08-01-15-16-36-0400__CAM_FRONT__1533151477612404_inv_depth_md4all.jpg} \hspace{1pt} &
        \includegraphics[width=0.19\textwidth]{figures/figures_supp/nuscenes/daytime/n008-2018-08-01-15-16-36-0400__CAM_FRONT__1533151477612404_inv_depth_ours.jpg} \\
\end{tabular}

    % 添加总体标题
    \caption{Qualitative examples in nuScenes \cite{nuscenes2019} daytime condition. Ours is comparable with previous state-of-the-art in daytime, while much better in adverse conditons.} 
    \label{fig:qualitative_nuscenes_daytime}
    % \vspace{-0.2cm}
\end{figure*}

\begin{figure*}[!t]
    \centering
    % 第一部分: nuScenes
    \begin{tabular}{c @{} c @{} c  @{} c @{} c @{} c }
        &
        \includegraphics[width=0.19\textwidth]{figures/figures_supp/nuscenes/nighttime/n015-2018-11-21-19-21-35+0800__CAM_FRONT__1542799328662460.jpg} 
        \hspace{1pt} &
        \includegraphics[width=0.19\textwidth]{figures/figures_supp/nuscenes/nighttime/n015-2018-11-21-19-21-35+0800__CAM_FRONT__1542799328662460_depth_gt.jpg} \hspace{1pt} &
        \includegraphics[width=0.19\textwidth]{figures/figures_supp/nuscenes/nighttime/n015-2018-11-21-19-21-35+0800__CAM_FRONT__1542799328662460_inv_depth_robust_depth.jpg} \hspace{1pt} &
        \includegraphics[width=0.19\textwidth]{figures/figures_supp/nuscenes/nighttime/n015-2018-11-21-19-21-35+0800__CAM_FRONT__1542799328662460_inv_depth_md4all.jpg} \hspace{1pt} &
        \includegraphics[width=0.19\textwidth]{figures/figures_supp/nuscenes/nighttime/n015-2018-11-21-19-21-35+0800__CAM_FRONT__1542799328662460_inv_depth_ours.jpg} \\
        
        %  row 2
        &
        \includegraphics[width=0.19\textwidth]{figures/figures_supp/nuscenes/nighttime/n015-2018-11-21-19-21-35+0800__CAM_FRONT__1542799343662460.jpg} \hspace{1pt} &
        \includegraphics[width=0.19\textwidth]{figures/figures_supp/nuscenes/nighttime/n015-2018-11-21-19-21-35+0800__CAM_FRONT__1542799343662460_depth_gt.jpg} \hspace{1pt} &
        \includegraphics[width=0.19\textwidth]{figures/figures_supp/nuscenes/nighttime/n015-2018-11-21-19-21-35+0800__CAM_FRONT__1542799343662460_inv_depth_robust_depth.jpg} \hspace{1pt} &
        \includegraphics[width=0.19\textwidth]{figures/figures_supp/nuscenes/nighttime/n015-2018-11-21-19-21-35+0800__CAM_FRONT__1542799343662460_inv_depth_md4all.jpg} \hspace{1pt} &
        \includegraphics[width=0.19\textwidth]{figures/figures_supp/nuscenes/nighttime/n015-2018-11-21-19-21-35+0800__CAM_FRONT__1542799343662460_inv_depth_ours.jpg} \\
\end{tabular}

    % 添加总体标题
    \caption{Qualitative examples in nuScenes \cite{nuscenes2019} nighttime condition. The challenging parts that our method is capable of predicting are emphasized by green boxes.} 
    \label{fig:qualitative_nuscenes_nighttime}
    % \vspace{-0.2cm}
\end{figure*}

\begin{figure*}[!t]
    \centering
    % 第一部分: nuScenes
    \begin{tabular}{c @{} c @{} c  @{} c @{} c @{} c }
        &
        \includegraphics[width=0.19\textwidth]{figures/figures_supp/nuscenes/rain/n008-2018-09-18-14-35-12-0400__CAM_FRONT__1537295950362404.jpg} \hspace{1pt} &
        \includegraphics[width=0.19\textwidth]{figures/figures_supp/nuscenes/rain/n008-2018-09-18-14-35-12-0400__CAM_FRONT__1537295950362404_depth_gt.jpg} \hspace{1pt} &
        \includegraphics[width=0.19\textwidth]{figures/figures_supp/nuscenes/rain/n008-2018-09-18-14-35-12-0400__CAM_FRONT__1537295950362404_inv_depth_robust_depth.jpg} \hspace{1pt} &
        \includegraphics[width=0.19\textwidth]{figures/figures_supp/nuscenes/rain/n008-2018-09-18-14-35-12-0400__CAM_FRONT__1537295950362404_inv_depth_md4all.jpg} \hspace{1pt} &
        \includegraphics[width=0.19\textwidth]{figures/figures_supp/nuscenes/rain/n008-2018-09-18-14-35-12-0400__CAM_FRONT__1537295950362404_inv_depth_ours.jpg} \\
        
        %  row 2
        &
        \includegraphics[width=0.19\textwidth]{figures/figures_supp/nuscenes/rain/n008-2018-09-18-14-35-12-0400__CAM_FRONT__1537296186362404.jpg} \hspace{1pt} &
        \includegraphics[width=0.19\textwidth]{figures/figures_supp/nuscenes/rain/n008-2018-09-18-14-35-12-0400__CAM_FRONT__1537296186362404_depth_gt.jpg} \hspace{1pt} &
        \includegraphics[width=0.19\textwidth]{figures/figures_supp/nuscenes/rain/n008-2018-09-18-14-35-12-0400__CAM_FRONT__1537296186362404_inv_depth_robust_depth.jpg} \hspace{1pt} &
        \includegraphics[width=0.19\textwidth]{figures/figures_supp/nuscenes/rain/n008-2018-09-18-14-35-12-0400__CAM_FRONT__1537296186362404_inv_depth_md4all.jpg} \hspace{1pt} &
        \includegraphics[width=0.19\textwidth]{figures/figures_supp/nuscenes/rain/n008-2018-09-18-14-35-12-0400__CAM_FRONT__1537296186362404_inv_depth_ours.jpg} \\
\end{tabular}

    % 添加总体标题
    \caption{Qualitative examples in nuScenes \cite{nuscenes2019} rain condition. The challenging parts that our method is capable of predicting are emphasized by green boxes.} 
    \label{fig:qualitative_nuscenes_rain}
    % \vspace{-0.2cm}
\end{figure*}

\begin{figure*}[!t]
    \centering
    % 第一部分: robotcar
    \begin{tabular}{c @{} c @{} c  @{} c @{} c @{} c }
        &
        \includegraphics[width=0.24\textwidth]{figures/figures_supp/robotcar/1418133821107956.jpg} \hspace{1pt} &
        \includegraphics[width=0.24\textwidth]{figures/figures_supp/robotcar/1418133821107956_depth_gt.jpg} \hspace{1pt} &
        \includegraphics[width=0.24\textwidth]{figures/figures_supp/robotcar/1418133821107956_inv_depth_md4all.jpg} \hspace{1pt} &
        \includegraphics[width=0.24\textwidth]{figures/figures_supp/robotcar/1418133821107956_inv_depth_ours.jpg}  \\
        
        %  row 2
        &
        \includegraphics[width=0.24\textwidth]{figures/figures_supp/robotcar/1418134251236839.jpg} \hspace{1pt} &
        \includegraphics[width=0.24\textwidth]{figures/figures_supp/robotcar/1418134251236839_depth_gt.jpg} \hspace{1pt} &
        \includegraphics[width=0.24\textwidth]{figures/figures_supp/robotcar/1418134251236839_inv_depth_md4all.jpg} \hspace{1pt} &
        \includegraphics[width=0.24\textwidth]{figures/figures_supp/robotcar/1418134251236839_inv_depth_ours.jpg}  \\

        %  row 3
        &
        \includegraphics[width=0.24\textwidth]{figures/figures_supp/robotcar/1418756851280111.jpg} \hspace{1pt} &
        \includegraphics[width=0.24\textwidth]{figures/figures_supp/robotcar/1418756851280111_depth_gt.jpg} \hspace{1pt} &
        \includegraphics[width=0.24\textwidth]{figures/figures_supp/robotcar/1418756851280111_inv_depth_md4all.jpg} \hspace{1pt} &
        \includegraphics[width=0.24\textwidth]{figures/figures_supp/robotcar/1418756851280111_inv_depth_ours.jpg}  \\
\end{tabular}

    % 添加总体标题
    \caption{Qualitative examples in Robotcar \cite{robotcar} daytime-nighttime conditions. The challenging parts that our method is capable of predicting are emphasized by green boxes.}
    \label{fig:qualitative_robotcar}
    % \vspace{-0.2cm}
\end{figure*}

\begin{figure*}[!t]
    \centering
    % 第一部分: DrivingStereo
    \begin{tabular}{c @{} c @{} c  @{} c @{} c @{} c }
        &
        \includegraphics[width=0.19\textwidth,height=0.105\textwidth]{figures/figures_supp/drivingstereo/rain/2018-08-17-09-45-58_2018-08-17-10-11-18-951.jpg} \hspace{1pt} &
        \includegraphics[width=0.19\textwidth, height=0.105\textwidth]{figures/figures_supp/drivingstereo/rain/2018-08-17-09-45-58_2018-08-17-10-11-18-951_depth_gt.jpg} \hspace{1pt} &
        \includegraphics[width=0.19\textwidth]{figures/figures_supp/drivingstereo/rain/2018-08-17-09-45-58_2018-08-17-10-11-18-951_inv_depth_robust_depth.jpg} \hspace{1pt} &
        \includegraphics[width=0.19\textwidth]{figures/figures_supp/drivingstereo/rain/2018-08-17-09-45-58_2018-08-17-10-11-18-951_inv_depth_md4all.jpg} \hspace{1pt} &
        \includegraphics[width=0.19\textwidth]{figures/figures_supp/drivingstereo/rain/2018-08-17-09-45-58_2018-08-17-10-11-18-951_inv_depth_ours.jpg} \\
        
        %  row 2
        &
        \includegraphics[width=0.19\textwidth,height=0.105\textwidth]{figures/figures_supp/drivingstereo/rain/2018-08-17-09-45-58_2018-08-17-10-11-28-264.jpg} \hspace{1pt} &
        \includegraphics[width=0.19\textwidth, height=0.105\textwidth]{figures/figures_supp/drivingstereo/rain/2018-08-17-09-45-58_2018-08-17-10-11-28-264_depth_gt.jpg} \hspace{1pt} &
        \includegraphics[width=0.19\textwidth]{figures/figures_supp/drivingstereo/rain/2018-08-17-09-45-58_2018-08-17-10-11-28-264_inv_depth_robust_depth.jpg} \hspace{1pt} &
        \includegraphics[width=0.19\textwidth]{figures/figures_supp/drivingstereo/rain/2018-08-17-09-45-58_2018-08-17-10-11-28-264_inv_depth_md4all.jpg} \hspace{1pt} &
        \includegraphics[width=0.19\textwidth]{figures/figures_supp/drivingstereo/rain/2018-08-17-09-45-58_2018-08-17-10-11-28-264_inv_depth_ours.jpg} \\
\end{tabular}

    % 添加总体标题
    \caption{Qualitative examples in DrivingStereo \cite{drivingstereo} rain condition. The challenging parts that our method is capable of predicting are emphasized by green boxes.} 
    \label{fig:qualitative_drivingstereo_rain}
    % \vspace{-0.2cm}
\end{figure*}

\begin{figure*}[!t]
    \centering
    % 第一部分: DrivingStereo
    \begin{tabular}{c @{} c @{} c  @{} c @{} c @{} c }
        &
        \includegraphics[width=0.19\textwidth,height=0.105\textwidth]{figures/figures_supp/drivingstereo/fog/2018-10-25-07-37-26_2018-10-25-07-42-52-291.jpg} \hspace{1pt} &
        \includegraphics[width=0.19\textwidth, height=0.105\textwidth]{figures/figures_supp/drivingstereo/fog/2018-10-25-07-37-26_2018-10-25-07-42-52-291_depth_gt.jpg} \hspace{1pt} &
        \includegraphics[width=0.19\textwidth]{figures/figures_supp/drivingstereo/fog/2018-10-25-07-37-26_2018-10-25-07-42-52-291_inv_depth_robust_depth.jpg} \hspace{1pt} &
        \includegraphics[width=0.19\textwidth]{figures/figures_supp/drivingstereo/fog/2018-10-25-07-37-26_2018-10-25-07-42-52-291_inv_depth_md4all.jpg} \hspace{1pt} &
        \includegraphics[width=0.19\textwidth]{figures/figures_supp/drivingstereo/fog/2018-10-25-07-37-26_2018-10-25-07-42-52-291_inv_depth_ours.jpg} \\
        
        %  row 2
        &
        \includegraphics[width=0.19\textwidth,height=0.105\textwidth]{figures/figures_supp/drivingstereo/fog/2018-10-25-07-37-26_2018-10-25-07-43-11-208.jpg} \hspace{1pt} &
        \includegraphics[width=0.19\textwidth, height=0.105\textwidth]{figures/figures_supp/drivingstereo/fog/2018-10-25-07-37-26_2018-10-25-07-43-11-208_depth_gt.jpg} \hspace{1pt} &
        \includegraphics[width=0.19\textwidth]{figures/figures_supp/drivingstereo/fog/2018-10-25-07-37-26_2018-10-25-07-43-11-208_inv_depth_robust_depth.jpg} \hspace{1pt} &
        \includegraphics[width=0.19\textwidth]{figures/figures_supp/drivingstereo/fog/2018-10-25-07-37-26_2018-10-25-07-43-11-208_inv_depth_md4all.jpg} \hspace{1pt} &
        \includegraphics[width=0.19\textwidth]{figures/figures_supp/drivingstereo/fog/2018-10-25-07-37-26_2018-10-25-07-43-11-208_inv_depth_ours.jpg} \\
\end{tabular}

    % 添加总体标题
    \caption{Qualitative examples in DrivingStereo \cite{drivingstereo} fog condition. The challenging parts that our method is capable of predicting are emphasized by green boxes.} 
    \label{fig:qualitative_drivingstereo_fog}
    % \vspace{-0.2cm}
\end{figure*}

\begin{figure*}[!t]
    \centering
    % 第一部分: DrivingStereo
    \begin{tabular}{c @{} c @{} c  @{} c @{} c @{} c }
        &
        \includegraphics[width=0.24\textwidth]{figures/figures_supp/viz_consistency/night/0_daytimerain_img_0.png} \hspace{1pt} &
        \includegraphics[width=0.24\textwidth]{figures/figures_supp/viz_consistency/night/0_dual_teacher_consistency.jpg} \hspace{1pt} &
        \includegraphics[width=0.24\textwidth]{figures/figures_supp/viz_consistency/night/0_daytimeteacher_inv_depth.jpg} \hspace{1pt} &
        \includegraphics[width=0.24\textwidth]{figures/figures_supp/viz_consistency/night/0_teacher_inv_depth.jpg} \hspace{1pt} \\
        
        %  row 2
        &
        \includegraphics[width=0.24\textwidth]{figures/figures_supp/viz_consistency/night/1_daytimerain_img_0.png} \hspace{1pt} &
        \includegraphics[width=0.24\textwidth]{figures/figures_supp/viz_consistency/night/1_dual_teacher_consistency.jpg} \hspace{1pt} &
        \includegraphics[width=0.24\textwidth]{figures/figures_supp/viz_consistency/night/1_daytimeteacher_inv_depth.jpg} \hspace{1pt} &
        \includegraphics[width=0.24\textwidth]{figures/figures_supp/viz_consistency/night/1_teacher_inv_depth.jpg} \hspace{1pt} \\

        %  row 3
        &
        \includegraphics[width=0.24\textwidth]{figures/figures_supp/viz_consistency/night/2_daytimerain_img_0.png} \hspace{1pt} &
        \includegraphics[width=0.24\textwidth]{figures/figures_supp/viz_consistency/night/2_dual_teacher_consistency.jpg} \hspace{1pt} &
        \includegraphics[width=0.24\textwidth]{figures/figures_supp/viz_consistency/night/2_daytimeteacher_inv_depth.jpg} \hspace{1pt} &
        \includegraphics[width=0.24\textwidth]{figures/figures_supp/viz_consistency/night/2_teacher_inv_depth.jpg} \hspace{1pt} \\
\end{tabular}

    % 添加总体标题
    \caption{Visualization of the consistency maps in nighttime and rain conditions.} 
    \label{fig:viz_consistency}
    % \vspace{-0.2cm}
\end{figure*}

\clearpage
\clearpage  
\newpage
{
    
    \small
    \bibliographystyle{ieeenat_fullname}
    \bibliography{main}
}

% WARNING: do not forget to delete the supplementary pages from your submission 
% \input{sec/X_suppl}